\documentclass[]{style}

\usepackage{xargs}

\usepackage{multirow}

\usepackage{cleveref}

\usepackage{amsmath}
\usepackage{dsfont}
\usepackage{array}
\usepackage{longtable}
\usepackage{xurl}
\usepackage{needspace}
\usepackage{tikz}
\usetikzlibrary{arrows.meta,calc,positioning,shapes.geometric}
\usepackage{pgfplots}
\pgfplotsset{compat=1.18}
\usepgfplotslibrary{groupplots}

\usepackage{mathrsfs}
\usepackage{adjustbox}
\usepackage{multirow}
\usepackage{multicol}
\usepackage{tcolorbox}
\usepackage{changepage}
\usepackage{enumitem}
\usepackage{graphicx}
\usepackage{amssymb}
\usepackage{xcolor}
\usepackage{float}
\usepackage{multirow}
\usepackage{threeparttable}
\usepackage{graphicx}
\usepackage{wrapfig}
\usepackage[table]{xcolor}  %
\usepackage{colortbl}       %
\usepackage{tabularx} %
\usepackage{makecell} %
\usepackage[dvipsnames]{xcolor}

\newcolumntype{g}{>{\columncolor{gray!10}}c} %
\newcolumntype{A}{>{\raggedright\arraybackslash}p{4.5cm}} %
\newcolumntype{Y}{>{\raggedright\arraybackslash}X} %

\definecolor{catgray}{gray}{0.9}
\definecolor{skyblue}{rgb}{0.53,0.81,0.92} %

\colorlet{skyblue!30}{skyblue!30!white} %

\colorlet{themeboxborder}{WeChatDeepGreen}

\newtcolorbox{evolbox}[2][]{%
  enhanced,
  colframe=themeboxborder,
  colback=white,
  coltitle=white,
  rounded corners,
  boxrule=1pt,
  titlerule=0pt,
  toptitle=1mm,
  bottomtitle=1mm,
  fonttitle=\bfseries,
  width=#2\textwidth, %
  #1
}

\usepackage{url}

\PassOptionsToPackage{table,xcdraw}{xcolor}
\usepackage{titletoc}
\usepackage{placeins}
\usepackage{pifont}

\colorlet{RowTheme}{WeChatMint}
\definecolor{RowRed}{HTML}{F9EAEA}
\definecolor{Top1}{HTML}{50DB4B} %
\definecolor{Top2}{HTML}{A5FFA2} %
\definecolor{Top3}{HTML}{D9FFD9} %
\definecolor{Sub1}{HTML}{EAB8B8}
\definecolor{Sub2}{HTML}{E4E4E4}
\newcommand{\cmark}{\textcolor{green!60!black}{\checkmark}}
\newcommand{\xmark}{\textcolor{red!70!black}{\ding{55}}}

\definecolor{gearred}{HTML}{D85140}
\definecolor{reprablue}{HTML}{5384ED}
\definecolor{steorange}{HTML}{EF8444}
\definecolor{softgreen}{HTML}{658E40}

\renewcommand{\emph}[1]{\textit{#1}}

\definecolor{codepink}{RGB}{220,20,120}
\definecolor{codegreen}{RGB}{0,150,0}
\definecolor{codegray}{RGB}{140,140,140}
\definecolor{codeorange}{RGB}{230,120,60}

\lstdefinestyle{pytorchstyle}{
    language=Python,
    basicstyle=\ttfamily\small,
    keywordstyle=\color{codepink}\bfseries,
    commentstyle=\color{codegray}\itshape,
    stringstyle=\color{codeorange},
    numberstyle=\tiny\color{codegray},
    numbers=none,
    showstringspaces=false,
    breaklines=true,
    frame=none,
    columns=fullflexible,
    keepspaces=true,
    xleftmargin=1.5em
}

\providecommand{\todo}[1]{\textcolor{WeChatDanger}{\textbf{[TODO] #1}}}

\newcommand{\modelname}{WeVisForensics\xspace}

\newcommand{\ReportAuthorNames}{%
  Yipeng Du\textsuperscript{1,2,*},\quad Yuxi Li\textsuperscript{2,\ensuremath{\dagger}},\quad
  Ying Zhang\textsuperscript{2},\quad Chong Sun\textsuperscript{2},\\[3pt]
  Chen Li\textsuperscript{2,\ensuremath{\ddagger}},\quad Jing Lyu\textsuperscript{2},\quad
  Ying Tai\textsuperscript{1,\ensuremath{\ddagger}}%
}

\newcommand{\ReportContributionNote}{%
  \textsuperscript{*}Work done as an intern at WeChat.\enspace
  \textsuperscript{\ensuremath{\dagger}}Project leader.\enspace
  \textsuperscript{\ensuremath{\ddagger}}Corresponding authors.%
}

\newcommand{\ReportProjectPageURL}{https://tencent.github.io/WeVisForensics}
\newcommand{\ReportGitHubURL}{https://github.com/Tencent/WeVisForensics}
\newcommand{\ReportHuggingFaceURL}{https://huggingface.co/tencent/WeVisForensics}

\hypersetup{
  pdftitle={WeVisForensics: Self-Evolving Agent for Omni AI-Generated Video Detection},
  pdfsubject={Agentic AI-generated video detection},
  pdfkeywords={AI-generated video detection, Multimodal LLM, Agentic AI, Self-evolving}
}

\usepackage{microtype}
\usepackage{multirow}
\usepackage{colortbl}
\usepackage[font=small]{subcaption}
\usepackage[most]{tcolorbox}
\definecolor{tablecategory}{RGB}{239,244,248}

\newcommand{\KL}{D_{\mathrm{KL}}}

\newcommand{\methodname}{WeVisForensics}
\newcommand{\datasetname}{ForenCue-30K}
\newcommand{\benchname}{AgenticCue}
\newcommand{\trainname}{DTPO}

\renewcommand{\methodname}{\modelname}
\definecolor{cvprblue}{rgb}{0.21,0.49,0.74}
\newcommand{\pub}[1]{{\color{gray}{\tiny{[{#1}]}}}}

\colorlet{OursRowColor}{WeChatGreen!12}
\newcommand{\oursrow}{\rowcolor{OursRowColor}}
\newcolumntype{C}{>{\centering\arraybackslash}X}
\newcommand{\ExpTableTextBold}[1]{{\fontsize{8.3}{9}\selectfont\bfseries\sffamily #1}}
\newcommand{\ExpTableSetup}{%
  \small
  \let\TableTextBold\ExpTableTextBold
  \setlength{\tabcolsep}{3pt}%
  \setlength{\aboverulesep}{0pt}%
  \setlength{\belowrulesep}{0pt}%
  \renewcommand{\arraystretch}{1.2}}

\usepackage{xspace}
\usepackage{environ}
\usepackage{adjustbox}

\AtBeginDocument{%
}
\let\ReportOrigResizebox\resizebox
\renewcommand{\resizebox}[3]{%
  \def\ReportResizeHeight{#2}%
  \def\ReportResizeBang{!}%
  \ifx\ReportResizeHeight\ReportResizeBang
    \adjustbox{max width=\textwidth}{#3}%
  \else
    \ReportOrigResizebox{#1}{#2}{#3}%
  \fi
}

\let\reportsection\section
\renewcommand{\section}{\Needspace*{5\baselineskip}\reportsection}

\newcommand\blfootnote[1]{%
  \begingroup
  \renewcommand\thefootnote{}%
  \begin{NoHyper}%
    \footnotetext{#1}%
    \end{NoHyper}%
  \endgroup
}
\title{\modelname{}: Self-Evolving Agent for\\
Omni AI-Generated Video Detection}
\author{\ReportAuthorNames}
\affiliation[1]{Nanjing University}
\affiliation[2]{WeChat Vision, Tencent Inc.}

\abstract{\ReportAbstract}
\let\abstract\relax

\NewEnviron{abstract}{\global\let\ReportAbstract\BODY}
\begin{abstract}
\fontsize{9pt}{10.6pt}\selectfont
Recent advances in video generation call for reliable detection grounded in explicit forensic evidence. However, current methods primarily depend on single-pass chain-of-thought reasoning that overlooks brief or localized artifacts, or on predefined multi-agent workflows that offer limited flexibility in selecting relevant inspections. Even with adaptive investigation, optimizing the investigation policy alone does not directly improve the forensic capabilities of fixed experts. We introduce \textbf{\methodname{}}, an agentic framework in which a main investigator agent coordinates a self-evolving library of forensic inspector sub-agents spanning visual and audio cues. The investigator learns which artifact type to examine, over which window, and when to conclude, and these interactions provide shared feedback for both levels of agents. Our dual-timescale policy optimization strategy, \textbf{\trainname{}}, closes this loop through fast reinforcement learning of the investigator and slower refinement of the inspectors, guided by annotated artifact intervals, to report subtle artifacts they once dismissed. Furthermore, we curate \textbf{\datasetname{}}, combining role-specific forensic annotations and interaction trajectories to supervise individual inspections and their coordination, with an accompanying reinforcement-learning pool. Experiments show that \methodname{} averages 95.2\% detection accuracy across in-domain, out-of-domain and agentic benchmarks, 9.0\% above the strongest prior detector, while grounding its verdicts in localized evidence.
\end{abstract}

\DeclareRobustCommand{\frontlinkicon}[1]{%
  \raisebox{-0.3em}{\includegraphics[width=0.025\linewidth]{#1}}}
\DeclareRobustCommand{\frontresourcelink}[2]{%
  \ifstrempty{#1}{\todo{#2}}{\href{#1}{\texttt{#1}}}}
\checkdata[\frontlinkicon{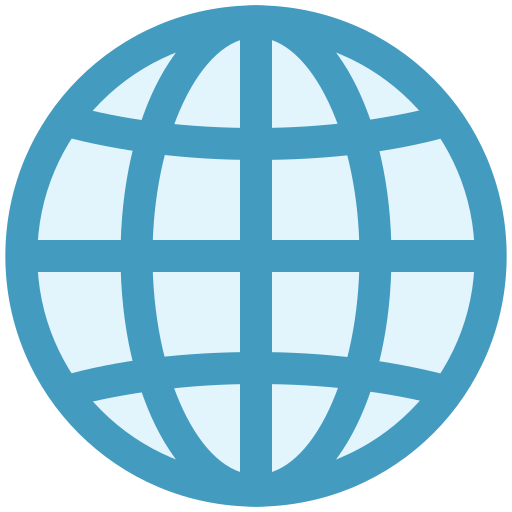}~~Project Page]{%
  \frontresourcelink{\ReportProjectPageURL}{Project page URL}}
\checkdata[\frontlinkicon{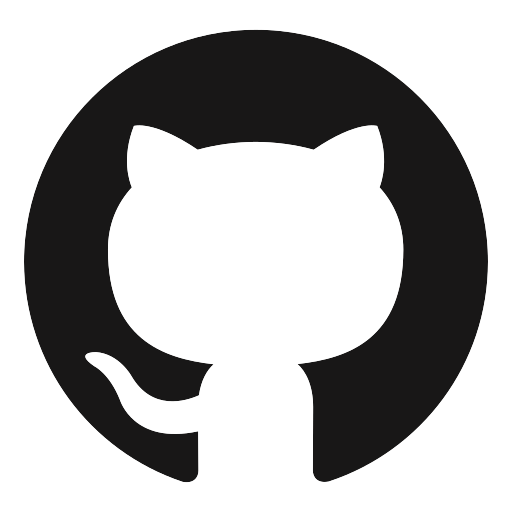}~~Code]{%
  \frontresourcelink{\ReportGitHubURL}{GitHub URL}}
\checkdata[\frontlinkicon{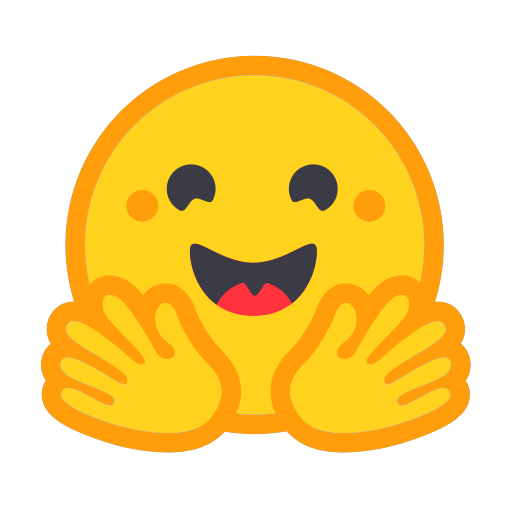}~~Weights]{%
  \frontresourcelink{\ReportHuggingFaceURL}{Hugging Face URL}}
\titleteaser{\noindent\begin{minipage}{\linewidth}
  \centering
  \includegraphics[width=\linewidth,height=2.93in,keepaspectratio]{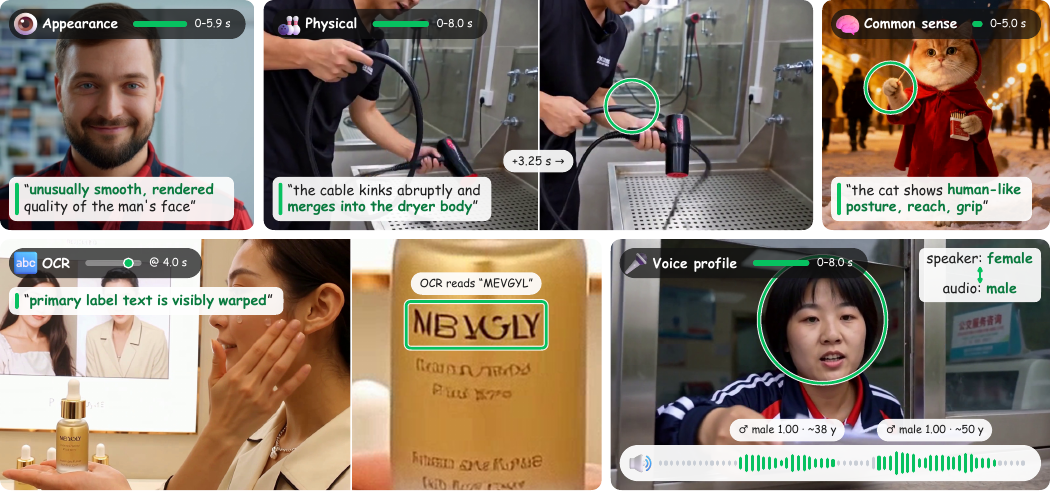}
  \captionsetup{hypcap=false,skip=1pt}
  \captionof{figure}{\small\textbf{Overview of \methodname{}'s role-specific forensic evidence across appearance, physics, common sense, on-screen text, and audio.}}
  \label{fig:teaser}
\end{minipage}
}

\begin{document}

\enlargethispage{14pt}
\blfootnote{\ReportContributionNote}
\maketitle

\begin{figure*}[!t]
  \centering
  \includegraphics[width=\textwidth]{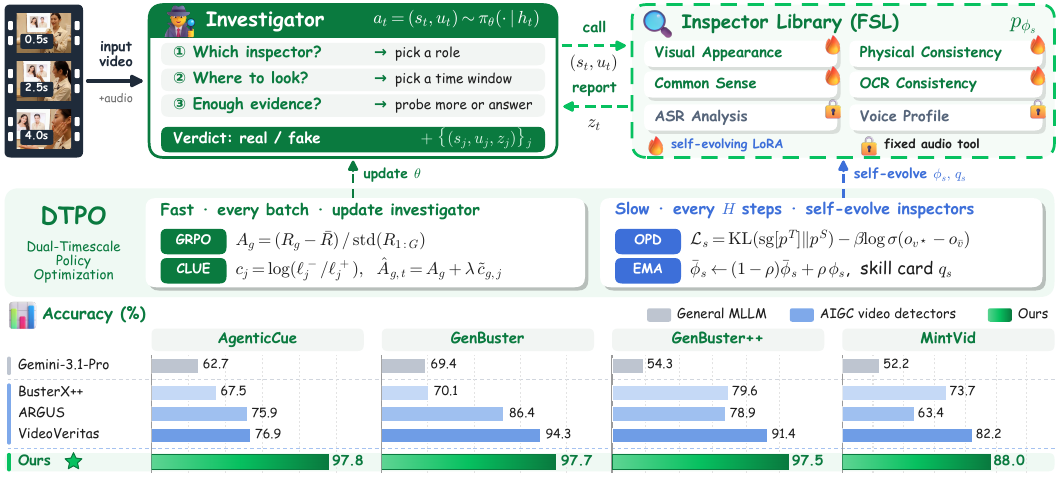}
  \caption{\textbf{Overview of \methodname{}.} \emph{Top:} the investigator $\pi_\theta$ repeatedly calls an inspector $s_t$ on a temporal scope $u_t$, reads its report $z_t$, and issues a verdict with the evidence chain. \emph{Middle:} \trainname{} updates the investigator on every batch, while the trainable inspectors self-evolve on a slower timescale. \emph{Bottom:} accuracy on the four evaluation benchmarks against representative detectors.}
  \label{fig:overview}
\end{figure*}

\section{Introduction}
\label{sec:intro}
Rapid advances in video generation have improved the visual fidelity, temporal coherence, and realism of synthetic content, bringing new possibilities to media creation. The same progress makes fabricated recordings increasingly difficult to distinguish from authentic ones, raising concerns over misinformation, impersonation, and public trust. Reliable AI-generated video detection is therefore becoming an essential component of media forensics, where decisive cues span visual, textual, and audio dimensions and call for role-specific inspection (Figure~\ref{fig:teaser}).

To meet this need, recent studies~\citep{wen2025busterx,wen2025busterx++,park2026vidguard,li2026skyra,tan2026videoveritas,jiang2026ivy,shen2026omnivl,liu2026vidforensics,li2026omni,shen2026omnivlpro,zou2026multi,qi2026dvar,wu2026safeguard} use multimodal large language models (MLLMs) either to reason about authenticity in natural language or to coordinate expert analysis. However, as illustrated in Figure~\ref{fig:compare}, these approaches still face two challenges in how forensic evidence is examined and used. \emph{a) Holistic single-pass reasoning.} CoT-centric methods~\citep{tan2026videoveritas,li2026skyra,park2026vidguard,wen2025busterx,wen2025busterx++} judge the whole video in one pass, with one model covering every artifact type. Artifacts confined to a brief interval or a small region are easily missed or explained away by hallucinated rationales (Figure~\ref{fig:motivation}a), and claims entangled in one free-form trace cannot be checked or supervised individually. \emph{b) Static workflows or coordinator-only learning.} Agent-based detectors either run a fixed expert pipeline~\citep{wu2026safeguard,qi2026dvar,lihermes} or train only the coordinator over fixed specialists. ARGUS~\citep{zou2026multi} collects four expert reports and freezes the observers while training the judge, so irrelevant reports can outweigh the one that finds the artifact (Figure~\ref{fig:motivation}b). Tool-augmented detectors~\citep{shen2026omnivlpro,wu2026tue} instead learn when to invoke tools, yet keep tool backends fixed. Both remain bounded by the local accuracy of frozen sub-agents, since an inspector that misses a subtle cue limits what better routing or aggregation can recover (Figure~\ref{fig:motivation}, right).

\begin{figure*}[t]
\centering
\includegraphics[width=.98\textwidth]{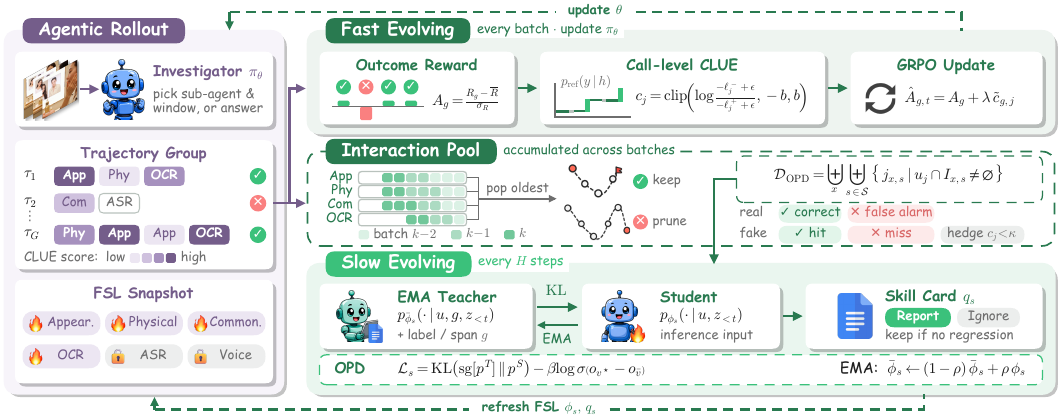}
\caption{\textbf{\methodname{} and its training strategy, \trainname{}.} The investigator rolls out with a fixed FSL snapshot and is updated on every batch with outcome rewards and CLUE. Grounded calls ($\mathcal D_{\mathrm{OPD}}$) queue by role in the interaction pool, from which the slow process distills privileged EMA teachers into students, revises skill cards, and refreshes the FSL every $H$ steps.}
\label{fig:method}
\end{figure*}

To overcome these challenges, we introduce \textbf{\methodname{}} (Figure~\ref{fig:overview}), an agentic paradigm that couples adaptive evidence collection with a self-evolving library of forensic inspector sub-agents coordinated by a main investigator agent along two complementary designs. \emph{1) Adaptive forensic analysis.} The Forensic Sub-agent Library (FSL) holds trainable inspectors that examine visual appearance, physical consistency, commonsense plausibility, and on-screen text, while audio tools inspect the soundtrack when one is present. Each call specifies a forensic role and a temporal scope, and the selected inspector returns findings that inform subsequent reasoning. The investigator uses these observations to request further inspections or issue an authenticity verdict, so evidence collection adapts to each video beyond a predefined workflow. \emph{2) Investigator-inspector self-evolution.} Adaptive dispatch still relies on what each inspector can detect, so the inspectors must also improve from the calls they serve. Yet because the investigator decides which inputs the inspectors encounter while their responses shape its decisions, updating both on every batch makes each a moving target for the other. Our \textbf{\trainname{}} (\textbf{D}ual-\textbf{T}imescale \textbf{P}olicy \textbf{O}ptimization) therefore trains both from shared trajectories on two timescales (Figure~\ref{fig:method}). The investigator's inspection policy is updated on every rollout batch using outcome rewards and call-level feedback. The inspectors accumulate the calls issued by the investigator, keep those grounded in annotated artifact intervals, select responses with interaction feedback for privileged distillation, revise their skill cards of reporting rules, and periodically refresh the FSL. The refined inspectors then return to subsequent rollouts, so the agent evolves from its own environment.

Moreover, existing forensic video datasets~\citep{wen2025busterx,jiang2026ivy,li2026skyra} supervise a single detector with a verdict or a holistic rationale, providing neither role-specific evidence for sub-agents nor trajectories for coordinating them. We therefore curate \textbf{\datasetname{}}, which decomposes forensic supervision by role. Each record isolates one forensic perspective and specifies a verdict, a confidence, and the temporal interval of the artifact, so that a sub-agent learns to report what its role can verify, including when the evidence is inconclusive. Agentic trajectories are assembled from executed sub-agent calls rather than imagined observations, so every decision is conditioned on evidence the main agent would actually receive. An accompanying RL pool supports \trainname{}. Since existing benchmarks score only the final verdict, we build \textbf{\benchname{}}, an agentic benchmark that groups generated videos by forensic category and evaluates both detection accuracy and whether an agent invokes the relevant sub-agents.
Our contributions are threefold:
\begin{itemize}[leftmargin=*,itemsep=2pt,topsep=3pt]
\item We propose \textbf{\trainname{}}, an RL algorithm in which the investigator and its inspectors co-evolve from shared interactions on fast and slow timescales. Inspectors refine their weights and skill cards on the very calls the investigator issues, so its rollouts become their curriculum.
\item We present \textbf{\methodname{}}, an agentic video forensics framework in which a main agent adaptively dispatches inspections to the FSL. The invocation strategy and sub-agent competence are jointly learned through \trainname{}.
\item We curate \textbf{\datasetname{}}, a role-structured dataset with agentic trajectories covering diverse generators, and introduce \textbf{\benchname{}}, the first agentic forensic benchmark that evaluates both accuracy and call precision.
\end{itemize}

\section{\methodname{} Agent Framework}
\label{sec:method}
\methodname{} treats AI-generated video detection as an investigation rather than a single act of perception. A human forensic analyst rarely judges a video from one glance: they form a hypothesis from the overall content, zoom in on the moment and the region where it can be tested, consult the appropriate specialist, and keep collecting evidence until the conclusion is supported. \methodname{} turns this workflow into an agent system with two levels. A main agent, the \emph{investigator}, plans the investigation, and a Forensic Sub-agent Library (FSL) of \emph{inspectors} examines the video from complementary forensic perspectives and returns structured evidence.

The framework is built on three design principles.
\textbf{(i) Separation of planning and perception.} The investigator decides \emph{what} to check and \emph{where}, while each inspector decides \emph{whether} the artifact of its specialty is present in the requested scope. Planning thus scales with the complexity of a video, and perception remains specialized and verifiable.
\textbf{(ii) Role-decomposed evidence.} Every inspection returns a verdict, a calibrated confidence, and a visibly grounded explanation for a single forensic perspective. Each piece of evidence can therefore be audited, supervised, and credited on its own, instead of being entangled in one free-form rationale.
\textbf{(iii) Evolvable components.} The parts of the system whose competence depends on the changing distribution of generators, namely the investigation policy and the visual inspectors, are trained to improve from the system's own interactions, whereas measurement-like tools stay fixed and provide a stable reference. Sec.~\ref{sec:protocol} specifies the investigation protocol, Sec.~\ref{sec:pool} describes each member of the FSL, and Sec.~\ref{sec:training} presents the training pipeline.

\subsection{Investigation Protocol}
\label{sec:protocol}
\paragraph{Observation and actions.}
An investigation starts from a global overview of the video: frames sampled at approximately 1~FPS across the full duration and uniformly reduced to at most 16, each tagged with its absolute timestamp, together with the duration and an indicator of whether an audio stream is present. At every turn, the investigator emits a single structured message consisting of a reasoning field and an action. The reasoning states the current hypothesis, what the evidence gathered so far does or does not establish, and why the next inspection is informative. Before the final answer, it synthesizes all returned evidence, including contrary and inconclusive findings. The action either calls one tool with its arguments, a temporal window $[t_s,t_e]$ for the window-based inspectors, a single timestamp for OCR, and no arguments for the whole-track audio tools, or terminates the investigation with a real/fake verdict.

Formally, a video $x$ with label $y\in\{\text{real},\text{fake}\}$ and duration $T_x$ is first summarized by the overview
\begin{equation}
 o_x=\big(\{(f_k,\nu_k)\}_{k=1}^{K},\;T_x,\;\alpha_x\big),
\end{equation}
where $f_k$ is the frame at absolute timestamp $\nu_k$ and $\alpha_x\in\{0,1\}$ indicates an audio stream. Let $\pi_\theta$ denote the main policy and $\{p_{\phi_s}\}_{s\in\mathcal S}$ the FSL members. At turn $t$, the history $h_t=(o_x,a_1,z_1,\ldots,a_{t-1},z_{t-1})$ contains the overview and all preceding interactions. An inspection action $a_t=(s_t,u_t)$ specifies the sub-agent $s_t$ and its scope $u_t$:
\begin{equation}
 a_t\sim\pi_\theta(\cdot\mid h_t),\qquad
 z_t\sim p_{\phi_{s_t}}(\cdot\mid v(x,u_t),q_{s_t}),
\end{equation}
where $v(x,u_t)$ is the input extracted from scope $u_t$ (sampled frames for visual sub-agents and the soundtrack for audio tools), $q_{s_t}$ is the role instruction together with the sub-agent's skill card (Sec.~\ref{sec:dualpo}), and $z_t$ is the response appended to the history. The investigation ends with an answer action $a_L=\textsc{answer}(\hat y)$, producing the trajectory $\tau=(o_x,a_1,z_1,\ldots,a_{L-1},z_{L-1},a_L)$.

\paragraph{Call budget and protocol rules.}
Let $C_t$ be the number of tool calls in $h_t$ and $\mathcal S_{\mathrm{aud}}\subset\mathcal S$ the audio tools. An inspection action $a_t=(s,u)$ is admissible only if
\begin{equation}
 C_t<4,\qquad (s,u)\notin\{a_{t'}\}_{t'<t},\qquad u\subseteq[0,T_x],\qquad s\in\mathcal S_{\mathrm{aud}}\Rightarrow\alpha_x=1,
 \label{eq:admissible}
\end{equation}
that is, at most four calls, no repeated call, scopes inside the video, and no audio tools on silent videos. The answer action is admissible once at least one call has succeeded, so an investigation has at most five assistant turns. A trajectory is \emph{valid} if every message parses and every action is admissible, and it is scored as
\begin{equation}
 R(\tau)=\mathbb 1[\tau\text{ is valid}]\cdot\mathbb 1[\hat y=y],
 \label{eq:correct}
\end{equation}
so malformed messages and rule violations count as errors. Within these limits, the investigator is instructed to choose the smallest sufficient set of inspections rather than exhausting the budget: an obvious, well-localized artifact often needs one call, and a further call is justified only by an independent uncertainty or complementary evidence. Temporal scopes are kept within one continuous shot. For videos longer than 30 seconds, the investigator targets a localized event, normally within a 15-second window, instead of a near-full-video range.

\paragraph{How reports are interpreted.}
Inspector reports are dimensional and may conflict. Each visual report carries one of three verdicts, \texttt{issue}, \texttt{no\_issue}, or \texttt{inconclusive}, with a confidence of high, medium, or low. The investigator treats \texttt{inconclusive} as moderate evidence that records a genuine limitation, never as a hidden decision in either direction, and treats \texttt{no\_text}, tool failures, and purely descriptive audio profiles as neutral unless they contain a concrete inconsistency. A \texttt{no\_issue} finding in one scope is not sufficient to establish the authenticity of the whole video, so the final verdict must integrate evidence across calls.

\subsection{Forensic Sub-agent Library}
\label{sec:pool}
The FSL covers six forensic perspectives (Table~\ref{tab:fsl}). Four visual inspectors, one each for appearance, physical consistency, common sense, and on-screen text, share the main agent's architecture but keep separate parameters, so that each can specialize without interfering with the others or with the investigation policy. Two audio tools analyze the soundtrack when one is present. Every visual inspector returns a structured report
\begin{equation}
 z=(v,\,c,\,e),\qquad v\in\mathcal V_s,\quad c\in\{\text{high},\text{medium},\text{low}\},
 \label{eq:report}
\end{equation}
where $\mathcal V_s$ is the verdict set of role $s$ (Table~\ref{tab:fsl}), $c$ is the confidence, and $e$ is an explanation that cites timestamps inside the inspected scope.

\begin{table}[t]
\centering\small
\caption{\textbf{Members of the Forensic Sub-agent Library.} Scope is the argument chosen by the investigator. Evolving members update their LoRA adapter and skill card during \trainname{}, while fixed members provide a stable measurement reference.}
\label{tab:fsl}
\providecommand{\cmark}{\textcolor{ForestGreen}{\checkmark}}
\setlength{\tabcolsep}{3pt}
\resizebox{\columnwidth}{!}{%
\begin{tabular}{llllc}
\toprule
Member & Scope & Input & Verdicts & Evolves\\
\midrule
Visual appearance & window & frames, 2~FPS & issue / no\_issue / inconcl. & \cmark\\
Physical consistency & window & frames, 2~FPS & issue / no\_issue / inconcl. & \cmark\\
Common sense & window & frames, 2~FPS & issue / no\_issue / inconcl. & \cmark\\
OCR consistency & timestamp & frame + text crops & text\_ok / text\_suspicious / no\_text & \cmark\\
Voice profile & whole track & soundtrack & descriptive profile & --\\
ASR analysis & whole track & transcript + timings & timing anomaly / natural & --\\
\bottomrule
\end{tabular}}
\end{table}

\paragraph{Window-based visual inspectors.}
The appearance, physical, and commonsense inspectors receive the frames of the requested window decoded at 2~FPS, each labeled with its absolute timestamp,
\begin{equation}
 v\big(x,[t_s,t_e]\big)=\big\{(x(\nu),\nu)\;\big|\;\nu\in[t_s,t_e],\ 2\nu\in\mathbb Z\big\},
\end{equation}
where $x(\nu)$ is the frame at time $\nu$, and follow a role-specific checklist. The \emph{appearance} inspector checks texture stability under motion, the integrity of fine details and boundaries such as hair, fingers, teeth, and small objects, unexplained local changes in sharpness, ghosting or duplication, and the temporal and geometric consistency of lighting, shadows, and reflections. The \emph{physical} inspector checks biological and kinematic motion, structural integrity of bodies and rigid objects, object permanence across occlusion, contact and grasping, and the dynamics of fluids and deformable materials. The \emph{commonsense} inspector checks causality, contextual consistency of objects and roles, affordances, and event order. The investigator calls appearance and physical checks on most videos and reserves commonsense checks for overviews that suggest a specific semantic or causal anomaly, such as an animal performing a human task or an event whose outcome contradicts ordinary world knowledge.
All three inspectors share one reporting discipline. \texttt{issue} requires a concrete anomaly supported by the supplied frames, and \texttt{no\_issue} requires the relevant content to be sufficiently observable and consistent. \texttt{inconclusive} is returned when the required evidence is absent, not meaningfully exercised, or obscured by sampling, blur, occlusion, or resolution. Confidence is calibrated to the strength and coverage of the evidence: a largely static subject cannot support a high-confidence physical judgment, and one clean frame cannot support a high-confidence appearance judgment. Normal cuts, camera motion, motion blur, compression, stylization, and animation are explicitly not anomalies by themselves, which prevents the inspectors from equating post-production with generation.

\paragraph{OCR inspector.}
Video generators still struggle to render text: glyphs that start legible often fuse, break, drift, or melt as the clip proceeds, which makes on-screen text an independent and highly diagnostic forensic channel. Text, however, occupies a small fraction of the frame, and fine stroke structure is lost once the frame is resized for a vision-language model. The OCR inspector is therefore a three-stage pipeline operating on the single frame $x(\nu)$ at the timestamp $\nu$ chosen by the investigator.
\emph{(1) Localization.} PaddleOCR~\citep{cui2025paddleocr} detects and recognizes candidate text lines $\mathcal B=\mathrm{Det}(x(\nu))$, each with a normalized bounding box and detection and recognition confidences. A frame with $\mathcal B=\varnothing$ returns \texttt{no\_text}.
\emph{(2) Text-presence routing.} OCR detectors fire on textures, faces, and compression patterns. The inspector views the full frame with the candidate list, decides only whether genuine text exists, and selects informative regions $\mathcal R\subseteq\mathcal B$ with $|\mathcal R|\le 6$, including subtitles, signs, labels, license plates, UI elements, and text on background objects. If every candidate is a false positive, $\mathcal R=\varnothing$ and it returns \texttt{no\_text}.
\emph{(3) Crop-level glyph inspection.} Each region $b_i\in\mathcal R$ is cropped with a 15\% margin at a higher pixel budget than the full frame. The inspector judges every crop independently against the full-frame context, assigning a crop label $g_i\in\mathcal Q=\{$\emph{regular}, \emph{warped}, \emph{generation-degraded}, \emph{too blurry}, \emph{stylized or logo}, \emph{inconclusive}$\}$, and looks for fused or broken strokes, malformed character topology, inconsistent repeated characters, and drifting baselines. Given its proposed verdict $\hat v$, the reported verdict is
\begin{equation}
 v=\begin{cases}
 \texttt{inconclusive}, & \hat v=\texttt{text\_suspicious}\ \text{and}\ g_i\notin\mathcal Q_{\mathrm{gen}}\ \forall\, b_i\in\mathcal R,\\
 \hat v, & \text{otherwise},
 \end{cases}
 \label{eq:ocr_gate}
\end{equation}
where $\mathcal Q_{\mathrm{gen}}=\{$\emph{warped}, \emph{generation-degraded}$\}$. A text-level forgery claim thus always rests on at least one crop with visibly generated glyphs.
The investigator calls OCR whenever the overview contains captions, signage, packaging, handwriting, or other text, and targets the timestamp where the text looks least stable. Figure~\ref{fig:case_serum_ocr} in Appendix~\ref{sec:append_more_cases} shows a complete example, in which an appearance check first flags an unnaturally smooth facial surface and a targeted OCR call then confirms malformed glyphs on the enlarged crops.

\paragraph{Audio tools.}
Generated videos increasingly carry synthesized speech, and an inconsistency between what is heard and what is seen is often more conspicuous than any visual artifact. The \emph{voice profile} tool performs speaker diarization, separates overlapping speech, and estimates for each speaker an age, female/male/child probabilities, and arousal, dominance, and valence scores. The report is descriptive by design: a strong age or gender conflict with a clearly visible speaking face is useful supporting evidence, whereas dubbing, voice-over, and multiple speakers must be discounted. The \emph{ASR analysis} tool transcribes the soundtrack with word-level timestamps and asks a frozen language model to review only concrete transcript and timing anomalies, such as robotic, uniformly spaced segments, against natural patterns such as hesitations and breathing pauses. Both tools are invoked on videos with an audio stream, typically when a speaking face or a talking-head scene is visible.

\paragraph{Trainable and fixed members.}
The FSL splits into $\mathcal S=\mathcal S_{\mathrm{evo}}\cup\mathcal S_{\mathrm{fix}}$, where $\mathcal S_{\mathrm{evo}}$ contains the four visual inspectors and $\mathcal S_{\mathrm{fix}}$ the two audio tools. Each member $s\in\mathcal S_{\mathrm{evo}}$ evolves during training through two complementary channels: its LoRA adapter $\phi_s$ is distilled from the investigator's own calls, and its skill card, a short list of \texttt{Report} and \texttt{Ignore} rules appended to the role instruction $q_s$, is revised by a reflector model (Sec.~\ref{sec:dualpo}). Visual artifacts are the cues that shift most rapidly as generators improve, and the calls issued by the investigator expose exactly the inputs, scopes, and difficulty levels on which the inspectors are consulted. The audio tools and the PaddleOCR detector remain fixed: their outputs are measurements, such as speaker attributes, transcripts, and text boxes, whose meaning should not drift while the investigator learns to interpret them, and the scarcity of labeled audio forgeries would make their adaptation unreliable. The investigator itself is optimized with reinforcement learning, so the system improves both in what it asks and in what it is told.

\section{Data Construction}
\label{sec:data}

\begin{figure*}[t]
\centering
\includegraphics[width=.98\textwidth]{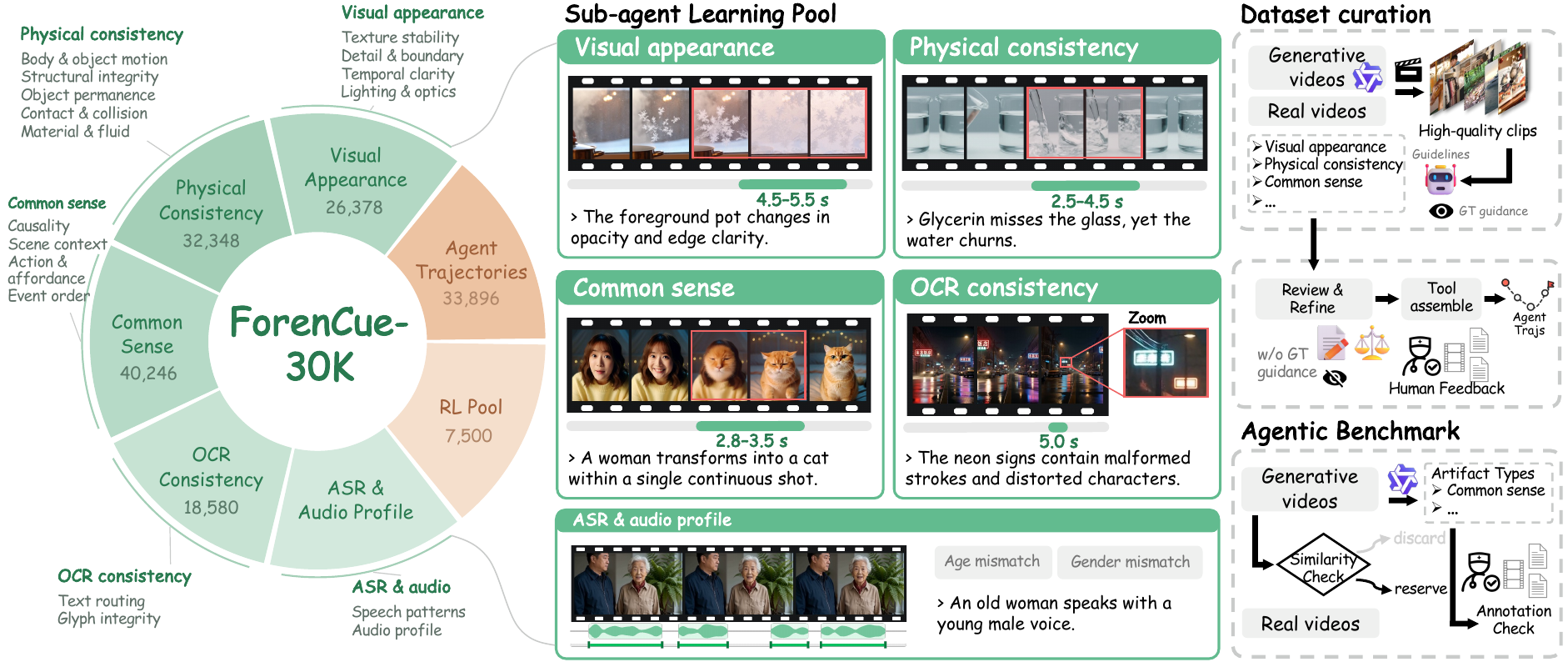}
\caption{\textbf{Overview of \datasetname{}.} Curated videos are annotated with role-specific artifact evidence and converted into sub-agent records and interaction trajectories, and an RL pool supports \trainname{}. Examples show artifacts targeted by each sub-agent.}
\label{fig:dataset}
\end{figure*}

Built from 30{,}000 curated videos, \textbf{\datasetname{}} provides multi-level supervision for our framework, comprising 117{,}552 role-specific records for the sub-agents, 33{,}896 interaction trajectories for the main agent, and a 7{,}500-row RL pool for \trainname{} (Figure~\ref{fig:dataset} and Table~\ref{tab:datacmp}). Its construction follows three principles: \textbf{broad generator coverage}, \textbf{a role-aligned forensic taxonomy}, and \textbf{temporally grounded evidence annotation}.

Generated videos are primarily curated from public datasets (GenBuster-200K~\citep{wen2025busterx}, RewardData~\citep{fu2025learning}, VBench-2.0~\citep{zheng2025vbench}, and AIGVDBench~\citep{ma2026your}) covering open-source and earlier-generation commercial models, supplemented with videos generated through AI video-generation platforms and specialized forgery domains including digital humans and speaker-mismatched audio forgeries. The generated training videos include CogVideoX~\citep{yang2025cogvideox}, HunyuanVideo~\citep{kong2024hunyuanvideo}, LTX-Video~\citep{hacohen2024ltx}, EasyAnimate~\citep{xu2024easyanimate}, Wan~2.1/2.2~\citep{wan2025wan}, Gen-2/3, Pika~1.5, Kling~1.0/1.5, MiniMax-Video-01, Veo~3, and Vidu~Q1. Real videos are sourced from OpenVid-1M~\citep{nan2025openvid} and Koala-36M~\citep{wang2025koala}, supplemented with videos collected from the stock media platforms Pexels and Pixabay.

\subsection{Role-Grounded Supervision Construction}
Supervision is constructed in a pipeline that mirrors the structure of the agent (Figure~\ref{fig:dataset}). Each curated video is first annotated separately for every forensic role, the annotations are then converted into training records for the inspectors, and executed inspector calls are finally assembled into investigation trajectories for the investigator. Every stage keeps the same role taxonomy, so the evidence an inspector learns to report is exactly the evidence the investigator learns to request and interpret.

\paragraph{Role-structured taxonomy.}
Instead of one holistic rationale per video, supervision is organized by forensic perspective and subdivided into fine-grained artifact dimensions. Matching the six FSL roles, we categorize artifacts into \emph{Visual appearance} (26{,}378 records), \emph{Physical consistency} (32{,}348), \emph{Common sense} (40{,}246), \emph{OCR consistency} (18{,}580), \emph{ASR analysis}, and \emph{Audio profile}. Visual appearance covers texture and detail integrity, boundary blending, ghosting, and the consistency of lighting, shadows, and reflections. Physical consistency covers body and object structure, kinematics, object permanence, contact, and fluid or deformable dynamics. Common sense covers causality, affordances, contextual roles, and event order, and OCR consistency covers glyph structure and the temporal stability of rendered text. The audio roles cover the plausibility of speech timing and the agreement between the voice and the visible speaker. Each record isolates one perspective and specifies a verdict (issue, no-issue, or inconclusive), a confidence, an explanation, and the temporal interval of the artifact. Real videos contribute records under the same format, so every role sees both outcomes. This balance teaches each inspector to report only what its role can verify, rather than to equate ``generated'' with ``an artifact of my type is visible''.

\paragraph{Evidence annotation.}
For each video and role, a rubric-conditioned VLM annotator examines the timestamped frames under a role-specific guideline listing the artifact dimensions above and source-provenance constraints, localizing observable artifacts within temporal intervals and writing frame-grounded explanations of what is visibly wrong. Explanations must refer strictly to observable visual evidence, and roles without visible anomalies are recorded as no-issue even on generated videos, preventing forced attribution. A cross-family VLM verifier then checks each candidate against the video and refines or rejects it, filtering hallucinated explanations, over-generalized commonsense claims, and imprecise intervals. Human reviewers inspect sampled annotations per source and role during prompt development and source onboarding, and their findings are fed back into the guidelines and filtering rules, while audio inconsistencies are annotated directly by reviewers who watch each clip with sound (Appendix~\ref{sec:human_review}). In a quantitative audit of 340 video--role annotations, $99.1\%$ are supported by the visual evidence.

\begin{table}[t]
\centering\small
\caption{\textbf{Comparison with existing forensic video training data.} Role: per-perspective records. Temp.: artifact intervals. Traj.: call--observation trajectories. Agentic: tool-oriented evaluation.}
\label{tab:datacmp}
\providecommand{\cmark}{\textcolor{ForestGreen}{\checkmark}}
\providecommand{\xmark}{\textcolor{BrickRed}{$\times$}}
\setlength{\tabcolsep}{3pt}
\resizebox{\columnwidth}{!}{%
\begin{tabular}{lcccccc}
\toprule
Dataset & Expl. & Role & Temp. & OCR/Aud. & Traj. & Agentic\\
\midrule
GenBuster-200K~\citep{wen2025busterx} & \xmark & \xmark & \xmark & \xmark & \xmark & \xmark\\
IVY-FAKE~\citep{jiang2026ivy} & \cmark & \xmark & \xmark & \xmark & \xmark & \xmark\\
ViF-CoT-4K~\citep{li2026skyra} & \cmark & \xmark & \cmark & \xmark & \xmark & \xmark\\
FaceVid-Forensics-100K~\citep{zou2026multi} & \cmark & \cmark & \xmark & \xmark & \xmark & \xmark\\
\midrule
\oursrow \textbf{\datasetname{} (ours)} & \cmark & \cmark & \cmark & \cmark & \cmark & \cmark\\
\bottomrule
\end{tabular}}
\end{table}

\paragraph{Inspector records.}
Each annotation is converted into a record whose input is exactly what the inspector observes at inference: the frames of the requested window at 2~FPS for the window-based roles, and the frame at the requested timestamp for OCR. The target is the structured report, comprising the verdict, the confidence, and an explanation that cites timestamps within the window. Because the windows are specified in absolute video time, the same annotation can supervise the inspector and later serve as the grounding interval $I_{x,s}$ for selecting calls during \trainname{}.

\paragraph{Trajectory synthesis.}
Interaction trajectories are constructed through execution-grounded synthesis. Given the video overview and verified role evidence as training-only context, a VLM trajectory planner decides which roles and temporal scopes to consult, modeling how an expert investigation prioritizes relevant checks. Every planned call is then executed by the corresponding inspector, and the planner writes the reasoning of each turn conditioned on the reports actually returned, so that the investigator never learns from imagined observations. All training-only metadata are stripped from student-facing messages, conditioning each decision strictly on inference-time observations. Trajectories cover both real and generated videos, use about two calls on average under the four-call budget, and reach $96.0\%$ appropriate tool selection in a human audit of 100 trajectories.

\paragraph{RL pool.}
The name \datasetname{} refers to the core video supervision set, while expanded annotation records, trajectories, and the accompanying RL pool are counted separately (Figure~\ref{fig:dataset}). The RL pool contains 7{,}500 rows spanning 15 collections and 56 source subsets, with 3{,}750 unique real and 3{,}180 unique generated clips repeated with source-dependent weights to balance labels and sources. Each generated clip carries its role-specific artifact intervals, which ground the selection of inspector calls in the slow process, and each clip carries a verified set of relevant roles, which defines the tool-coverage reward (Sec.~\ref{sec:dualpo}).

\paragraph{Quality control.}
Export checks validate record structure, role-specific fields, video duration, and the availability of sampled frames. For interaction trajectories, construction checks match call identifiers to stored responses and enforce the action budget and the protocol rules of Sec.~\ref{sec:protocol}, and automated text checks screen the student-facing messages for training-only metadata leakage. Each trajectory record retains the video and source identifiers, sampled-frame timestamps, ordered messages, and target verdict.

\subsection{\benchname{} Benchmark}
Existing benchmarks score only the final verdict, so an agent that reaches the right answer through irrelevant inspections, or that cannot handle a particular class of artifacts, remains invisible. \benchname{} is designed to expose both. Unlike the training set, which is dominated by open-source and earlier commercial models, it targets newer-generation commercial releases (Veo~3.1, Kling~2.5/3.0, Seedance~1.5/2.0, Hailuo~2.3/H3, Wan~2.6, Sora~2, Jimeng, and HappyHorse~1.0) alongside held-out digital-human, face-swap, and voice-mismatch audio forgeries, and thus also measures generalization to generators unseen during training.

\paragraph{Category construction.}
The benchmark spans five forensic categories, \emph{Visual}, \emph{Physical}, \emph{OCR}, \emph{Common sense}, and \emph{Audio}, each containing exactly 500 generated videos verified to exhibit that category's artifact. Candidate clips are first annotated by model annotators and then confirmed by human reviewers. A clip may belong to several categories, and 170 clips do, giving 2{,}330 unique generated videos. All categories share the same 500 real videos, so each category forms a balanced set of 1{,}000 videos. None of its videos appear in \datasetname{}.

\paragraph{Metrics.}
For each category, we report binary detection accuracy on its balanced set, counting invalid or missing outputs as errors. To evaluate whether the agent consults the right specialist, we additionally report invocation precision, the fraction of generated videos on which the agent invokes tool $s$ that actually call for role $s$. The reference set of relevant roles for the four visual tools comes from dense annotations in which two cross-family VLM annotators independently mark every role worth inspecting on each generated video and a label-blind judge resolves disagreements, while for audio it is the Audio category. Table~\ref{tab:prec_counts} lists the resulting label counts and the base rates obtained by invoking a tool on every video. Precision is always read together with accuracy, which would drop if the agent skipped relevant inspections. Details are given in Appendix~\ref{sec:append_impl}.

\begin{table}[t]
\centering\small
\caption{\textbf{Label counts behind sub-agent invocation precision (Table~\ref{tab:toolprec}).} Eligible counts the generated videos scored for each role, $|G_s|$ the videos that call for role $s$, and Base the precision of invoking tool $s$ on every eligible video, which is also the precision of the fixed-workflow ARGUS on its two observer roles.}
\label{tab:prec_counts}
\setlength{\tabcolsep}{3pt}
\begin{tabular}{lccccc}
\toprule
 & Appear. & Phys. & OCR & Comm. & Audio\\
\midrule
Eligible & 2{,}328 & 2{,}328 & 2{,}328 & 2{,}328 & 2{,}330\\
$|G_s|$ & 2{,}121 & 1{,}829 & 1{,}146 & 558 & 500\\
Base (\%) & 91.1 & 78.6 & 49.2 & 24.0 & 21.5\\
\bottomrule
\end{tabular}
\end{table}

\section{Training Pipeline}
\label{sec:training}
\methodname{} is trained in two stages. Supervised fine-tuning (SFT) on \datasetname{} first teaches the investigator the investigation protocol and each inspector its role-specific reporting format (Sec.~\ref{sec:sft}). \trainname{} then lets the investigator and the inspectors improve together from their own interactions (Sec.~\ref{sec:dualpo}), and a scalable implementation keeps the joint training practical (Sec.~\ref{sec:scalable}). Every model is adapted with low-rank adapters on a shared backbone, so that the investigator and all inspectors can be loaded and served as one backbone plus a few lightweight adapters, and both stages update every model in the same parameterization.

\subsection{Stage I: Supervised Initialization}
\label{sec:sft}
All models share the Qwen3.5-4B~\citep{qwen3.5} backbone. The investigator and the four visual inspectors are trained as five separate rank-64 LoRA~\citep{hu2021lora} adapters ($\alpha{=}128$) on all linear layers of the language model, and the backbone, the vision encoder, and the vision-language projector are frozen throughout, so that adaptation concentrates on forensic reasoning rather than low-level visual features.

Each model is trained with the standard masked next-token objective. A training sample is a token sequence $y=(y_1,\ldots,y_{L_y})$ with a loss mask $m_{y,j}\in\{0,1\}$ that equals one on model-authored tokens and zero on instructions, overview frames, inspector reports, tool outputs, and padding. For a model with parameters $\psi$ trained on a dataset $\mathcal D$,
\begin{equation}
 \mathcal L_{\mathrm{SFT}}(\psi;\mathcal D)=-\frac{\sum_{y\in\mathcal D}\sum_{j=1}^{L_y} m_{y,j}\log p_{\psi}(y_j\mid y_{<j})}{\sum_{y\in\mathcal D}\sum_{j=1}^{L_y} m_{y,j}},
 \label{eq:sft}
\end{equation}
so the loss averages the negative log-likelihood over all supervised tokens. The investigator minimizes $\mathcal L_{\mathrm{SFT}}(\theta;\mathcal D_{\mathrm{inv}})$ on interaction trajectories, where the mask covers the reasoning, every action, and the final verdict, and each inspector $s$ minimizes $\mathcal L_{\mathrm{SFT}}(\phi_s;\mathcal D_s)$ on its role-specific records, where the mask covers the structured report of Eq.~\eqref{eq:report}.

\paragraph{Investigator.}
The investigator is trained on the 33{,}896 agentic trajectories of \datasetname{}, extended with trajectories on additional videos covering a broader range of content. Each trajectory contains every executed call and returned report together with the reasoning that motivates each step and the final verdict, and is packed into one multi-turn sample with a maximum length of 32{,}768 tokens. The adapter is trained with a learning rate of $2{\times}10^{-4}$ and a cosine schedule with 5\% warm-up in two consecutive single-epoch stages with different data orderings, and the stage-1 adapter is merged into the backbone before stage 2 begins.

\paragraph{Inspectors.}
Each visual inspector is trained on its own portion of the 117{,}552 role-specific records, extended with records on additional videos. Each record pairs a scoped input with a verdict, a confidence, and an evidence-grounded explanation. Real and generated videos are kept in every role, so that an inspector does not learn to report an issue by default. The inspectors follow the same adapter recipe as the investigator, with a maximum length of 32{,}768 tokens (16{,}384 for OCR, whose input is a single frame). The resulting system consists of a single backbone plus five small adapters, which makes it easy to store and serve.

\subsection{Stage II: \trainname{}}
\label{sec:dualpo}
\trainname{} makes interaction experience the link between learning to request evidence and refining the inspectors that provide it. The investigator determines which inputs the inspectors encounter, and the resulting reports influence both its subsequent decisions and which interactions are selected for inspector training. Updating both at the same time would make each a moving target for the other. The fast process therefore updates only the investigator on every rollout batch, while the slow process accumulates role-specific interaction queues over multiple batches and aligns the inspectors with a slowly changing EMA teacher~\citep{tarvainen2017mean}. The updated FSL then supplies observations for later rollouts, closing the loop. Video-level labels and artifact annotations are used only during training. At inference, the learned investigator and inspectors execute the protocol of Sec.~\ref{sec:protocol} with fixed parameters.

\paragraph{Fast process: investigator policy learning.}
For each input, we collect $G$ interaction trajectories $\{\tau_g\}_{g=1}^{G}$ with the current main policy and a fixed FSL snapshot. Each trajectory receives the correctness reward $R_g=R(\tau_g)$ of Eq.~\eqref{eq:correct}, so invalid trajectories receive zero. We use GRPO~\citep{shao2024deepseekmath} to normalize rewards within the group,
\begin{equation}
 A_g=\frac{R_g-\overline R}{\operatorname{std}(R_1,\ldots,R_G)+\delta},
\end{equation}
where $\delta$ stabilizes normalization. The terminal reward says whether the investigation succeeded, but not which of its inspections contributed. We therefore augment it with \textbf{CLUE} (\textbf{C}all-\textbf{L}evel \textbf{U}tility \textbf{E}stimation), a per-action credit signal that measures the change in a frozen reference model's support for the correct label after each inspection call. The reference scores the known label $y$ before and after the response to call $j$. With probabilities normalized over the real/fake alternatives, let $\ell_j^{\mp}=\log p_{\mathrm{ref}}(y\mid h_j^{\mp})$, where $h_j^-$ and $h_j^+$ are the histories before and after the response. The CLUE score is
\begin{equation}
 c_j=\operatorname{clip}\!\left(
 \log\frac{-\ell_j^-+\epsilon}{-\ell_j^++\epsilon},-b,b\right),
 \label{eq:credit}
\end{equation}
where $\epsilon$ smooths the ratio and $b$ bounds the score. A positive score means that the returned report moved the reference toward the correct label, and a negative score means that it misled the reference. Because the reference is distinct from both the investigator and the inspector teachers, the score measures the informativeness of the evidence itself rather than the policy's own belief. We center scores within each GRPO group. Writing $c_{g,j}$ for the score of call $j$ in trajectory $g$, $\tilde c_{g,j}=c_{g,j}-\frac{1}{|\mathcal G_j|}\sum_{g'\in\mathcal G_j}c_{g',j}$, where $\mathcal G_j$ is the set of rollouts in the group that issue at least $j$ calls. The token advantages become
\begin{equation}
 \widehat A_{g,t}=\begin{cases}
 A_g+\lambda\tilde c_{g,j}, & t\in\mathcal I_{g,j},\\
 A_g, & t\in\mathcal M_g\setminus\mathcal I_g,\\
 0, & t\notin\mathcal M_g,
 \end{cases}
 \label{eq:action-advantage}
\end{equation}
where $\mathcal M_g$ contains main-agent output tokens, $\mathcal I_{g,j}$ is the token span of inspection action $j$, $\mathcal I_g=\bigcup_j\mathcal I_{g,j}$, and $\lambda$ weights call feedback. A clipped policy-ratio objective uses $\widehat A_{g,t}$ to update the investigator. Final-answer tokens retain $A_g$, and inspector observations and padding are masked from the loss. Inspection choices thus receive feedback on the evidence they produced, while the final decision remains trained by its outcome.

\paragraph{Tool-coverage reward.}
Correctness alone tolerates investigations that reach the right verdict through an irrelevant inspection, and such shortcuts generalize poorly to videos whose decisive artifact lies elsewhere. Each training video therefore carries a verified set of relevant forensic roles, derived from its role-specific annotations, and each trajectory receives the F1 score between this set and the set of tools it actually invoked. The F1 score is centered within the GRPO group and added to $A_g$ with weight $0.3$ after the correctness normalization, so it ranks investigations that share the same outcome without overriding correctness.

\paragraph{Slow process: collecting interactions for refinement.}
Every call to a trainable inspector during rollouts is a candidate refinement example, so the main policy determines which inputs, temporal scopes, and difficulty levels the inspectors are refined on. Candidates are first grounded in annotations. Let $I_{x,s}$ be the annotated artifact intervals of role $s$ on a generated video $x$, or the whole video if $x$ is real, and $u_j$ the scope inspected by call $j$. The calls kept for refinement are
\begin{equation}
 \mathcal D_{\mathrm{OPD}}=\biguplus_{x}\biguplus_{s\in\mathcal S}\big\{\,j_{x,s}\mid u_j\cap I_{x,s}\neq\varnothing\,\big\}.
\end{equation}
Calls on generated videos without an overlapping annotation are excluded, since failing to find an artifact in such a scope is not an error of that role. We then partition each kept response by its verdict and the video label. On a real video, a reported issue is a \emph{false alarm} and a no-issue verdict is \emph{correct}. On a generated video, a call that reports no issue is a \emph{miss} and one that reports an issue is a \emph{hit}.

A correct verdict can still hinder subsequent forensic reasoning, for example an inconclusive or low-confidence report that leaves the investigator without usable evidence. The CLUE score in Eq.~\eqref{eq:credit} identifies such responses: a correct response is relabeled as a \emph{hedge} when $c_j<\kappa$ and its verdict is inconclusive or not stated with high confidence. False alarms, misses, and hedges are the responses whose decisions must change, while the remaining correct responses are retained so that refinement preserves decisions that are already right. The final trajectory need not be incorrect: a locally unhelpful report warrants refinement even when the investigator recovers the correct answer. All retained calls, rather than one call per prompt and role, enter role-specific queues. When a role is refreshed, its examples are arranged into batches with equal numbers of real and generated videos by subsampling the majority label.

\paragraph{Privileged inspector distillation.}
Each example supplies its inspector input $u_i$, collected response $z_i$, and privileged context $g_i$. For role $s$, an exponential-moving-average (EMA) teacher receives $g_i$, following the training-only supervision principle of privileged-information distillation~\citep{penaloza2026privileged}, while the student retains its inference-time input. For a real video, $g_i$ states the label. For a generated video, $g_i$ lists the annotated intervals that overlap the inspected scope within the role's specialty and instructs the teacher to locate the corresponding anomaly in the supplied frames, treating the pointer as a cue for where to look rather than as evidence. As in on-policy distillation~\citep{agarwal2024policy}, teacher and student distributions at a collected response prefix are
\begin{equation}
 p^T_{i,t}=p_{\bar\phi_s}(\cdot\mid u_i,g_i,z_{i,<t}),\quad
 p^S_{i,t}=p_{\phi_s}(\cdot\mid u_i,z_{i,<t}).
\end{equation}
Let $t_i^\star$ be the position where the response commits to its verdict, $v_i^\star$ the label-consistent verdict token (issue or text-suspicious for generated videos, no-issue or text-ok for real ones), and $\bar v_i$ its opposite. The per-role objective is
\begin{equation}
\begin{aligned}
 \mathcal L_s=\frac{1}{N}\sum_{i=1}^{N}\Big[&
 \frac{\sum_{t=1}^{|z_i|} w_{i,t}\,
 \KL\!\left(\operatorname{sg}[p^T_{i,t}]\,\|\,p^S_{i,t}\right)}{\sum_{t=1}^{|z_i|} w_{i,t}}\\
 &-\beta\log\frac{e^{o_{i}(v_i^\star)}}{e^{o_{i}(v_i^\star)}+e^{o_{i}(\bar v_i)}}\Big],
\end{aligned}
 \label{eq:distill}
\end{equation}
where $\operatorname{sg}$ stops teacher gradients, $w_{i,t}=\omega$ at decision-field tokens and $1$ elsewhere, $o_i(\cdot)$ denotes student logits at $t_i^\star$, and $\beta$ weights the label correction of the verdict. The correction contrasts only the two opposing verdicts, so it corrects the decision without suppressing the inconclusive option or other vocabulary entries. For OCR, the same objective is further grounded on double-checked \emph{text-ok} regions drawn from generated videos, so the inspector learns that regular text on a forged clip is not itself evidence of forgery. For responses whose decisions must change, the distillation term is dropped when the privileged teacher still assigns its highest probability at $t_i^\star$ to the student's original verdict, in which case only the label correction is applied. The teacher supplies token distributions at the student's collected prefixes, so no replacement trajectory has to be generated. The teacher tracks the student via $\bar\phi_s\leftarrow(1-\rho)\bar\phi_s+\rho\phi_s$, and the student learning rate decays linearly over the investigator's schedule, so the FSL settles as the inspection policy converges.

\paragraph{Skill-card evolution.}
Adapter updates change how an inspector perceives. Some failure modes, however, are better expressed as explicit reporting rules, for example that motion blur on fast-moving hands is not a structural defect, or that a soft object deforming under direct contact is not a physics violation. Alongside its adapter weights, each trainable inspector therefore carries a skill card~\citep{wang2023voyager}, a short list of \texttt{Report} and \texttt{Ignore} rules appended to its role instruction and seen by student and teacher. After each distillation round, a frozen reflector LLM reviews recent calls that the current inspector still judges wrongly, with their labels, and proposes rule edits. A candidate card replaces the current one only if it yields more correct verdicts on held-out recent calls without losing any correct verdict on either real or generated videos. This acceptance test makes card edits monotone on recent experience and prevents the reflector from trading false alarms for misses. Examples of evolved cards are given in Appendix~\ref{sec:append_skill_cards}.

\paragraph{Synchronizing fast and slow updates.}
Training alternates rollout collection, investigator optimization, interaction selection, and periodic FSL refinement. Each batch is collected with the current main policy and a fixed inspector snapshot. After the investigator update, selected interactions enter role-specific queues. Every $H$ investigator steps, a synchronization barrier checks whether each queue contains enough examples to update its inspector. Eligible students are distilled, their skill cards are revised, and the new FSL snapshot is published before rollout collection resumes, while roles with insufficient examples retain their current parameters and keep their queues for the next barrier. This schedule lets the investigator learn between FSL updates while each refinement stage draws on accumulated interaction experience.

\subsection{Scalable Implementation}
\label{sec:scalable}
A single \trainname{} step runs $32\times G$ multi-turn investigations, each invoking up to four inspectors on video inputs, so inspector serving and FSL refreshes dominate the cost of training. The investigator and the four visual inspectors are served by separate vLLM engines that share the training cluster, and the rollout workers execute investigations asynchronously, so that a slow audio or OCR call does not stall other trajectories in the batch. The audio tools and PaddleOCR run as independent services. At each synchronization barrier, the four inspector refreshes are packed into two parallel lanes of GPUs, one handling visual appearance followed by common sense and the other physical consistency followed by OCR, which balances the per-lane workload because the roles accumulate very different numbers of calls. Refreshed adapters are merged into the backbone weights and published to the inference engines as full checkpoints rather than hot-swapped as LoRA modules, which we found to be numerically more robust on our hardware. Skill-card revision runs after publication. With a refresh every 20 steps, the full joint schedule of a few hundred investigator steps completes in about a day on our cluster.

\subsection{Implementation Details}
\label{sec:impl}
\paragraph{Main models.}
The base Qwen3.5-4B serves as the frozen CLUE reference $p_{\mathrm{ref}}$. \trainname{} starts from the SFT adapters of Sec.~\ref{sec:sft} and keeps every model in the adapter parameterization. The fast process runs GRPO with $G{=}4$ rollouts per video, a batch of 32 videos, and at most 4 inspector calls per trajectory for one epoch over the RL pool. The investigator adapter is trained with a learning rate that warms up linearly to $1{\times}10^{-5}$ over the first 60 steps. CLUE uses $\epsilon{=}0.1$, $b{=}1$, and $\lambda{=}0.2$, the tool-coverage bonus has weight $0.3$, and the hedge threshold is $\kappa{=}{-}0.05$. In the slow process, each visual inspector adapter is trained every $H{=}20$ investigator steps on class-balanced batches of 32 grounded calls, with a learning rate decayed linearly from $3{\times}10^{-5}$ over training progress, $\omega{=}5$, $\beta{=}1$, and $\rho{=}0.001$. The RL pool used here extends the pool of Sec.~\ref{sec:data} with additional labeled videos.

\section{Experiments}
\label{sec:experiments}

\subsection{Evaluation Setup}
\paragraph{Evaluation protocols.}
GenBuster-Bench (GB)~\citep{wen2025busterx} serves as the in-domain benchmark. Out-of-domain evaluation uses the video track of GenBuster-Bench++ (GB++)~\citep{wen2025busterx++} and the General split of MintVid~\citep{tan2026videoveritas}. We report binary detection accuracy for GB, GB++, each of the six generator categories of MintVid General, and the five \benchname{} categories. SoTA binary detectors are trained on our dataset using their original settings.

\paragraph{Implementation details.}
The main agent and sub-agents share one Qwen3.5-4B~\citep{qwen3.5} backbone, and each carries its own rank-64 LoRA~\citep{hu2021lora} adapter on all linear layers of the language model, initialized from its SFT adapter. The fast process runs GRPO with $G{=}4$ rollouts per video, a batch of 32 videos, and at most 4 sub-agent calls per trajectory for one epoch over the RL pool. The main-agent learning rate warms up linearly to $1{\times}10^{-5}$ over the first 60 steps. In the slow process, each sub-agent adapter is trained on class-balanced batches of 32 selected interactions with an EMA rate $\rho{=}0.001$. The served adapters are refreshed every 20 steps, and the sub-agent learning rate decays linearly from $3{\times}10^{-5}$ over training progress.

\subsection{Main Results}
\paragraph{Evaluation on standard benchmarks.}
Table~\ref{tab:main} shows that \methodname{} attains the strongest detection performance on both the in-domain and the out-of-domain benchmarks. On the in-domain split, binary detectors stay near chance, whereas MLLM-based detectors benefit from domain supervision, and \methodname{} reaches 97.7\%, surpassing VideoVeritas by 3.4\% and ARGUS by 11.3\%. On the out-of-domain splits, generic MLLMs suffer from low fake recall regardless of scale, and CoT detectors drop sharply across unseen generators. \methodname{} ranks first on GenBuster++ at 97.5\% and on the six-generator MintVid mean at 88.0\%, improving over the SFT agent on every generator category, which demonstrates strong generalization to unseen generators.

\begin{table*}[t]
\centering\small
\caption{\textbf{Performance comparison on AI-generated video detection.} MintVid General is split into six generator categories under the official repeated-reference protocol. Each Agentic category pairs 500 generated videos with 500 real videos. Avg.\ averages GenBuster, GenBuster++, the MintVid mean, and the Agentic mean. Best and second-best results are bold and underlined.}
\label{tab:main}
\setlength{\tabcolsep}{3pt}
\resizebox{\textwidth}{!}{%
\begin{tabular}{l*{14}{c}}
\toprule
\multirow{2}{*}{\textbf{Method}} & \multirow{2}{*}{\textbf{ID}} & \multicolumn{7}{c}{\textbf{OOD}} & \multicolumn{5}{c}{\textbf{OOD-Agentic}} & \multirow{2}{*}{\textbf{Avg.}}\\
\cmidrule(lr){3-9}\cmidrule(lr){10-14}
& & GB++ & Hailuo & Jimeng & Kling & Seedance & Sora2 & Wan2.5 & Visual & Physical & OCR & Common & Audio & \\
\midrule
\rowcolor{tablecategory}
\multicolumn{15}{l}{\textit{Binary Detectors}}\\
NSG-VD~\citep{zhang2026physics}~\pub{NeurIPS'25} & 46.4 & 53.7 & 57.1 & 48.5 & 52.6 & 54.7 & 57.1 & 51.0 & 41.7 & 43.2 & 41.7 & 34.2 & 44.4 & 48.7\\
ReStraV~\citep{interno2026ai}~\pub{NeurIPS'25} & 51.9 & 66.9 & 59.5 & 56.8 & 55.6 & 64.0 & 50.8 & 54.9 & 55.5 & 54.6 & 51.6 & 47.0 & 51.8 & 57.0\\
\midrule
\rowcolor{tablecategory}
\multicolumn{15}{l}{\textit{Generic MLLMs}}\\
Qwen3.5-4B~\citep{qwen3.5} & 82.9 & 73.2 & 63.5 & 56.5 & 65.1 & 73.8 & 57.0 & 73.2 & 73.7 & 71.8 & 83.8 & 94.4 & 58.1 & 74.3\\
Qwen3.5-397B~\citep{qwen3.5} & 65.1 & 66.7 & 54.7 & 52.0 & 55.8 & 63.0 & 51.4 & 63.4 & 69.2 & 66.7 & 86.0 & 94.5 & 54.1 & 65.7\\
Qwen3-VL-8B~\citep{bai2025qwen3} & 44.8 & 55.2 & 52.0 & 50.4 & 50.5 & 63.0 & 52.6 & 55.9 & 61.7 & 58.6 & 72.2 & 91.9 & 47.8 & 55.1\\
Qwen3-Omni-30B-A3B~\citep{xu2025qwen3} & 50.7 & 55.2 & 51.7 & 51.9 & 52.9 & 61.4 & 54.6 & 56.3 & 57.5 & 54.8 & 70.6 & 85.6 & 43.1 & 55.8\\
Gemini-3.1-Pro~\citep{deepmind2026gemini31pro} & 69.4 & 54.3 & 55.1 & 54.4 & 54.6 & 45.3 & 48.6 & 54.9 & 61.8 & 61.6 & 63.4 & 63.4 & 63.1 & 59.6\\
\midrule
\rowcolor{tablecategory}
\multicolumn{15}{l}{\textit{MLLM-based video forgery detectors}}\\
BusterX++~\citep{wen2025busterx++}~\pub{NeurIPS'26} & 70.1 & 79.6 & 70.4 & 71.7 & 82.3 & 71.8 & 58.7 & \textbf{87.3} & 64.4 & 65.1 & 71.1 & 87.1 & 49.9 & 72.7\\
DeepTraceReward~\citep{fu2025learning}~\pub{COLM'26} & 76.4 & 58.8 & 64.5 & 64.1 & 63.9 & 56.5 & 59.3 & 65.5 & 55.5 & 54.2 & 51.9 & 55.4 & 53.5 & 62.9\\
Skyra-RL~\citep{li2026skyra}~\pub{CVPR'26} & 68.2 & 55.7 & 51.7 & 51.7 & 53.5 & 63.4 & 50.6 & 59.6 & 83.4 & 83.8 & 74.0 & 86.9 & 68.5 & 64.6\\
VideoVeritas~\citep{tan2026videoveritas}~\pub{ICML'26} & \underline{94.3} & 91.4 & \underline{85.8} & \underline{80.8} & \underline{86.0} & 86.6 & 67.6 & \underline{86.3} & 72.4 & 75.3 & 76.0 & 91.5 & 69.3 & 86.2\\
ARGUS~\citep{zou2026multi} & 86.4 & 78.9 & 61.8 & 56.5 & 65.1 & 67.6 & 56.1 & 73.5 & 73.3 & 71.1 & 80.4 & 90.5 & 64.2 & 76.2\\
\midrule
\rowcolor{tablecategory}
\multicolumn{15}{l}{\textit{Agentic investigation}}\\
\oursrow \methodname{}-SFT & 93.9 & \underline{94.2} & 83.8 & 77.9 & 82.1 & \underline{95.6} & \underline{75.9} & 82.5 & \underline{92.4} & \underline{94.2} & \underline{93.6} & \underline{96.0} & \underline{95.2} & \underline{91.3}\\
\oursrow \textbf{\methodname{}-\trainname{}} & \textbf{97.7} & \textbf{97.5} & \textbf{88.9} & \textbf{82.9} & \textbf{89.6} & \textbf{98.6} & \textbf{82.4} & 85.6 & \textbf{97.3} & \textbf{98.0} & \textbf{98.0} & \textbf{98.8} & \textbf{97.0} & \textbf{95.2}\\
\bottomrule
\end{tabular}}
\end{table*}

\paragraph{Evaluation on \benchname{}.}
The five categories of \benchname{} expose the blind spots of prior paradigms across artifact types. Generic MLLMs handle commonsense violations but miss fine-grained anomalies, as Qwen3.5-397B reaches 94.5\% on commonsense but only 54.1\% on audio. Lacking specialized forensic tools, CoT detectors also degrade sharply, with DeepTraceReward~\citep{fu2025learning} near chance at 51.9--55.5\% across all five categories and BusterX++~\citep{wen2025busterx++} at 49.9\% on audio. In contrast, \methodname{} stays within 97.0--98.8\% across all five categories and leads VideoVeritas~\citep{tan2026videoveritas} by 27.7\% on audio and 20.9\% on average. \trainname{} improves over the SFT agent on all five categories, most on visual appearance by 4.9\% and on OCR by 4.4\%, demonstrating balanced detection.

\begin{table}[htbp]
\centering\footnotesize
\caption{\textbf{Sub-agent invocation precision on \benchname{}} (\%). Fraction of the
generated videos on which a tool is invoked that are annotated with its category.
\#Calls is the average number of executed tool invocations per video.}
\label{tab:toolprec}
\setlength{\tabcolsep}{3pt}
\resizebox{\columnwidth}{!}{%
\begin{tabular}{lccccc|c}
\toprule
Method & Appear. & Phys. & OCR & Comm. & Audio & \#Calls\\
\midrule
\methodname{}-SFT & \textbf{98.6} & 94.0 & 72.8 & 77.1 & 84.1 & 2.50\\
\oursrow \textbf{\methodname{}-\trainname{}} & \textbf{98.6} & \textbf{94.6} & \textbf{95.4} & \textbf{99.7} & \textbf{94.3} & \textbf{2.19}\\
\bottomrule
\end{tabular}}
\end{table}

\subsection{Ablation Studies}
\providecommand{\gain}[1]{\,{\scriptsize\textcolor{ForestGreen}{$\uparrow$#1}}}
Table~\ref{tab:ablation} isolates the fast and slow components of \trainname{} and yields three observations. First, optimizing the main policy against a frozen library mainly fits the training distribution. GRPO gains 2.9\% on GB and 2.8\% on GB++ over SFT, but only 1.3\% on MintVid and 0.4\% on \benchname{}. Adding CLUE under the same frozen library moves the average by only 0.5\%, because static sub-agents remain an evidence bottleneck. Second, slow sub-agent evolution breaks this bottleneck and drives out-of-domain transfer, lifting the average by 1.6\% over GRPO, with gains of 2.3\% on MintVid and 2.7\% on \benchname{}. Third, action-level credit becomes effective once it is paired with an evolving library, adding a further 1.4\% on MintVid and reaching the best overall average of 95.2\%, ahead of SFT by 3.9\% and GRPO by 2.0\%. Publishing the refined library every rollout batch instead of every $H{=}20$ steps also hurts transfer, lowering every unseen generator by up to 4.8\% and the OOD mean by 2.0\% in Table~\ref{tab:cadence}. The two processes are thus complementary, provided the library is refreshed at a stable cadence.

\begin{table}[htbp]
\centering\footnotesize
\caption{\textbf{Component ablation.} All variants start from the same SFT agent and FSL, train on the same RL pool. Accuracy (\%) follows the protocol of Table~\ref{tab:main}. Agentic is the mean accuracy over the five \benchname{} categories. Avg.\ averages GB, GB++, MintVid, and Agentic.}
\label{tab:ablation}
\setlength{\tabcolsep}{3.5pt}
\resizebox{\columnwidth}{!}{%
\begin{tabular}{lcccccc}
\toprule
Main agent & FSL & GB & GB++ & MintVid & Agentic & Avg.\\
\midrule
SFT & SFT & 93.9 & 94.2 & 83.0 & 94.3 & 91.3\\
GRPO & frozen & 96.8 & 97.0 & 84.3 & 94.7 & 93.2\\
GRPO + CLUE & frozen & 96.5 & 96.9 & 84.5 & 96.8 & 93.7\\
\trainname{} w/o CLUE & evolving & 97.4 & \textbf{97.6} & 86.6 & 97.4 & 94.8\\
\oursrow \trainname{} & evolving & \textbf{97.7} & 97.5 & \textbf{88.0} & \textbf{97.8} & \textbf{95.2}\\
\bottomrule
\end{tabular}
}
\end{table}

\providecommand{\gu}[1]{\,{\scriptsize\textcolor{ForestGreen}{$\uparrow$#1}}}
\providecommand{\gd}[1]{\,{\scriptsize\textcolor{ForestGreen}{$\downarrow$#1}}}
\providecommand{\ru}[1]{\,{\scriptsize\textcolor{BrickRed}{$\uparrow$#1}}}
\providecommand{\rd}[1]{\,{\scriptsize\textcolor{BrickRed}{$\downarrow$#1}}}
\begin{table*}[t]
\centering
\begin{minipage}[c]{0.6\textwidth}
\centering\footnotesize
\setlength{\tabcolsep}{3pt}
\begin{tabular}{lccccccc}
\toprule
 & \multicolumn{7}{c}{OOD generator categories}\\
\cmidrule(lr){2-8}
Schedule & Hailuo & Jimeng & Kling & Seedance & Sora2 & Wan2.5 & Mean\\
\midrule
\oursrow \trainname{} ($H{=}20$) & \textbf{88.9} & \textbf{82.9} & \textbf{89.6} & \textbf{98.6} & \textbf{82.4} & \textbf{85.6} & \textbf{88.0}\\
Every step & 84.1 & 82.3 & 87.0 & 98.2 & 80.1 & 84.3 & 86.0\\
Staged & 85.5 & 81.5 & 84.9 & 97.8 & 80.2 & 83.9 & 85.6\\
\midrule
$\Delta$ vs.\ every step & \gu{4.8} & \gu{0.6} & \gu{2.6} & \gu{0.4} & \gu{2.3} & \gu{1.3} & \gu{2.0}\\
$\Delta$ vs.\ staged & \gu{3.4} & \gu{1.4} & \gu{4.7} & \gu{0.8} & \gu{2.2} & \gu{1.7} & \gu{2.4}\\
\bottomrule
\end{tabular}
\end{minipage}\hfill
\begin{minipage}[c]{0.38\textwidth}
\centering
\raisebox{-4.4pt}[\height][\depth]{\includegraphics[width=\linewidth]{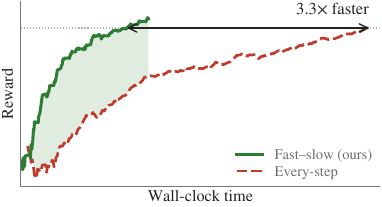}}
\end{minipage}
\caption{\textbf{Update cadence of the forensic library.} \emph{Left:} accuracy (\%) on unseen OOD generators comparing refreshing the library every $H{=}20$ steps as in \trainname{}, refreshing it on every rollout batch, and a staged schedule that first evolves the library and then freezes it while training the main agent. All start from the same SFT agent and library and are evaluated at the same step, and $\Delta$ is \trainname{} minus each alternative. \emph{Right:} training reward against wall-clock time.}
\label{tab:cadence}
\label{tab:cadence_full}
\end{table*}

\paragraph{Slow-fast training is effective.}
Table~\ref{tab:cadence} compares \trainname{} with two alternative schedules from the same SFT initialization. Refreshing the library on every rollout batch lowers accuracy on all six unseen OOD generators, with drops of 4.8\% on Hailuo, 2.6\% on Kling, and 2.3\% on Sora2, reducing the six-generator mean by 2.0\% despite similar training rewards. The staged schedule first evolves the library and then freezes it while training the main agent, so both levels improve but never adapt to each other. \trainname{} still leads it on every unseen generator, by 2.4\% on the six-generator mean, as co-evolution lifts fake recall across unseen generators from 70.2\% under the staged schedule and 71.3\% under every-step refresh to 74.8\% while preserving 100\% real-video recall. Strengthening the inspectors first or refreshing them on every batch thus fails to recover the out-of-domain transfer of two-timescale co-evolution.

\subsection{Does Sub-agent Evolution Improve Evidence?}
Video-level accuracy cannot separate a better inspection policy from better sub-agent responses. Table~\ref{tab:skills} therefore scores sub-agent calls against role-level annotations that mark which roles are worth inspecting on each test video and over which interval. Two cross-family VLM annotators label each video independently, and a label-blind judge resolves their disagreements. The evolving library sharpens the evidence of all three roles. The physical sub-agent improves on every benchmark, by 9.1--31.3\%, rising from 66.5\% to 97.8\% on GB as its inconclusive rate falls from 26.9\% to 0.3\%. On the unseen MintVid generators, appearance and physical F1 rise by 7.0\% and 19.2\%, and OCR evidence improves on every benchmark, by up to 11.8\% on GB++. A matched human review (Appendix~\ref{sec:human_review}) agrees, preferring the evolved responses in 65 pairs against 20 and cutting responses with any flaw from 30.5\% to 8.0\%. Unlike a frozen library, the evolving one thus improves the evidence itself, not only how the main agent uses it during investigation.

\providecommand{\gain}[1]{\,{\scriptsize\textcolor{ForestGreen}{$\uparrow$#1}}}
\providecommand{\drop}[1]{\,{\scriptsize\textcolor{BrickRed}{$\downarrow$#1}}}
\providecommand{\same}[1]{\,{\scriptsize\textcolor{gray}{$\pm$#1}}}
\begin{table}[htbp]
\centering\footnotesize
\caption{\textbf{Sub-agent evidence quality.} Sub-agent F1 (\%) against role-level annotations. A call counts when its role is annotated and its window overlaps the interval. Both libraries answer the same matched calls. Frozen: SFT library in GRPO. Evolving: refined by \trainname{}. OCR excludes text-presence routing.}
\label{tab:skills}
\setlength{\tabcolsep}{5pt}
\begin{tabular}{llcccc}
\toprule
Role & FSL & GB & GB++ & MintVid & \benchname{}\\
\midrule
\multirow{2}{*}{Appearance} & Frozen & 97.3 & 97.1 & 79.4 & 95.4\\
& \cellcolor{OursRowColor}Evolving & \cellcolor{OursRowColor}\textbf{98.1}\gain{0.8} & \cellcolor{OursRowColor}\textbf{97.5}\gain{0.4} & \cellcolor{OursRowColor}\textbf{86.4}\gain{7.0} & \cellcolor{OursRowColor}\textbf{98.7}\gain{3.3}\\
\midrule
\multirow{2}{*}{Physical} & Frozen & 66.5 & 85.1 & 72.7 & 89.4\\
& \cellcolor{OursRowColor}Evolving & \cellcolor{OursRowColor}\textbf{97.8}\gain{31.3} & \cellcolor{OursRowColor}\textbf{97.3}\gain{12.2} & \cellcolor{OursRowColor}\textbf{91.9}\gain{19.2} & \cellcolor{OursRowColor}\textbf{98.5}\gain{9.1}\\
\midrule
\multirow{2}{*}{OCR} & Frozen & 89.3 & 76.7 & 56.6 & 77.9\\
& \cellcolor{OursRowColor}Evolving & \cellcolor{OursRowColor}\textbf{92.7}\gain{3.4} & \cellcolor{OursRowColor}\textbf{88.5}\gain{11.8} & \cellcolor{OursRowColor}\textbf{58.9}\gain{2.3} & \cellcolor{OursRowColor}\textbf{84.0}\gain{6.1}\\
\bottomrule
\end{tabular}
\end{table}

Table~\ref{tab:toolprec} asks whether the main agent invokes the role a video calls for. Querying every tool on every video only attains the base rate, which limits ARGUS to 78.6\% on physical consistency. \methodname{} instead reaches 94.3--99.7\% precision on the physical, OCR, commonsense, and audio roles, 1.2--4.4 times their base rates, and \trainname{} improves physical precision by 0.6\% and both OCR and commonsense precision by 22.6\% over the SFT agent. Precision does not require exhaustive calls, as \trainname{} learns which inspector is informative and issues 12.4\% fewer calls than SFT yet gains accuracy on every \benchname{} category.

\begin{figure*}[t]
  \centering
  \includegraphics[width=.98\textwidth]{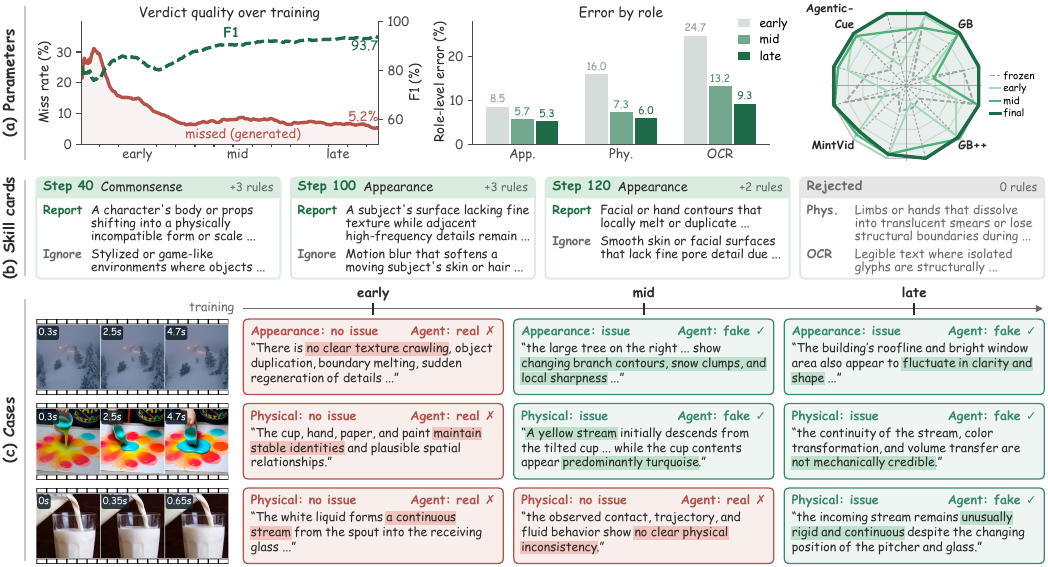}
  \caption{\textbf{Self-evolution of the forensic library over training.} \textbf{(a)} Sub-agent verdicts on training calls, scored against role-level annotations. Left: moving averages of the miss rate on generated videos and of the verdict F1 over LoRA refreshes (minor ticks). Middle: role-level error per sub-agent across the early, middle, and late phases of training. Right: evidence F1 of the frozen library and of the evolved library at the early, middle, and final checkpoints on four test benchmarks, each spanning the appearance, physical, and OCR axes clockwise. \textbf{(b)} Accepted edits, with one \texttt{Report} and one \texttt{Ignore} rule each. The gray card shows candidate rules that failed the acceptance test. \textbf{(c)} Three generated test videos inspected by the three checkpoints, with verbatim sub-agent excerpts and the agent's final answer.}
  \label{fig:visualevo}
\end{figure*}

\subsection{Visualizing Self-Evolution}
\begin{figure*}[t]
  \centering
  \includegraphics[width=\textwidth]{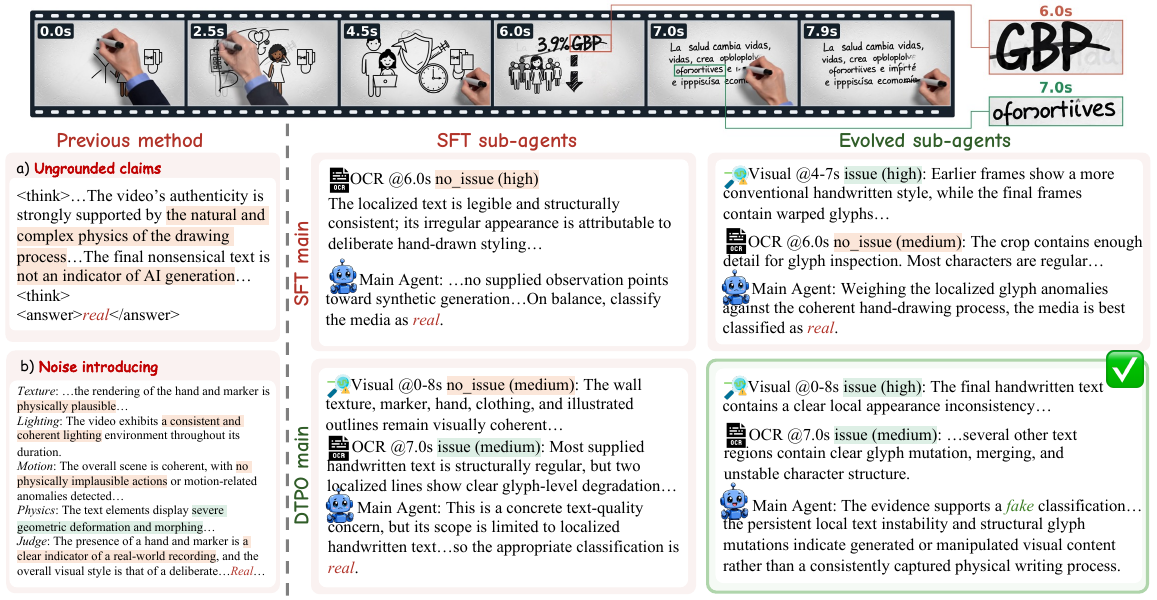}
  \caption{\textbf{Qualitative comparison on a generated video with warping handwritten text.} \emph{Left:} \emph{a)} The CoT-based VideoVeritas~\citep{tan2026videoveritas} dismisses the garbled text as stylistic without inspecting it. \emph{b)} In the fixed-workflow ARGUS~\citep{zou2026multi}, texture, lighting, and motion reports outweigh the detected text deformation. \emph{Right:} The SFT or \trainname{}-trained main agent paired with the SFT or evolved sub-agents. The evolved sub-agents flag the glyph anomalies, and only the trained main agent, which moves its OCR inspection to the warped text at 7\,s, turns them into a fake verdict.}
  \label{fig:motivation}
\end{figure*}

\paragraph{Both levels contribute.}
Table~\ref{tab:swap_avg} crosses the SFT and \trainname{}-trained main agents with the SFT and evolved libraries, the pairings illustrated in Figure~\ref{fig:motivation}. Evolving the library alone lifts the SFT investigator from 91.3\% to 92.0\%, and training the investigator alone reaches 93.2\%. Evolving both reaches 95.2\%, a gain of 3.9\% that exceeds the sum of the two single-level gains by 1.3\%. The two levels are therefore complementary: the trained investigator issues more targeted calls, and the evolved inspectors make each of those calls more informative.

\begin{table}[htbp]
\centering\footnotesize
\caption{\textbf{Contribution of each level to investigator--inspector co-evolution.} The investigator and inspectors before (SFT) or after (\trainname{}) co-evolution are paired in all four combinations, reporting average accuracy (\%) over GB, GB++, MintVid, and \benchname{}. Evolving either level alone helps, and evolving both gives the largest gain.}
\label{tab:swap_avg}
\begin{tabular}{lcc}
\toprule
Investigator $\backslash$ Inspectors & SFT & \trainname{}\\
\midrule
SFT & 91.3 & 92.0\\
\trainname{} & 93.2 & \textbf{95.2}\\
\bottomrule
\end{tabular}
\end{table}

Figure~\ref{fig:visualevo} traces the forensic library through training at three levels. At the parameter level, successive LoRA refreshes cut the missed anomalies on generated videos from 22.1\% to 5.2\% while false alarms on real videos also decline and verdict F1 rises from 80.3\% to 93.7\%, and every sub-agent errs progressively less across the early, middle, and late phases, with the OCR error falling from 24.7\% to 9.3\%. On the test benchmarks, the final library delivers higher evidence F1 than the frozen one along all three axes. At the prompt level, accepted skill-card edits add reporting rules to the appearance and commonsense sub-agents. At the case level, the same inspection returns different evidence over training. The early sub-agent calls each scene coherent and the agent answers real, whereas the final sub-agent names the regenerating texture, the color-shifting paint stream, or the rigid milk stream, and the agent answers fake. Figure~\ref{fig:motivation} shows the same effect at the agent level, where only the evolved sub-agents confirm the artifact. The gains thus come from sub-agents that learn to report artifacts they once called coherent.

\paragraph{A qualitative example.}
Figure~\ref{fig:motivation} follows one generated video whose handwritten text warps as the clip proceeds. VideoVeritas dismisses the garbled text as stylistic without inspecting it, and in ARGUS the texture, lighting, and motion reports outweigh the detected text deformation. Within \methodname{}, the SFT sub-agents judge the text structurally regular under either main agent. The evolved sub-agents flag its glyph anomalies, yet the SFT main agent still attributes them to hand-drawn styling and answers real. Only the \trainname{}-trained main agent, which targets its OCR inspection at the warped text at 7\,s, combines this evidence into a fake verdict, so the correction requires both levels to evolve. A complete multi-turn investigation before and after evolution and further comparisons are given in Appendix~\ref{sec:append_qual_comparisons}, and a full trajectory in Section~\ref{sec:append_cases}.

\def\casepart{1}\colorlet{tracegreen}{WeChatDeepGreen}
\colorlet{tracebackground}{white}
\definecolor{verdictfake}{RGB}{176,48,48}
\definecolor{verdictreal}{RGB}{40,130,70}
\definecolor{flagred}{RGB}{176,48,48}
\definecolor{flaggreen}{RGB}{40,130,70}
\definecolor{flagneutral}{RGB}{110,110,110}

\providecommand{\casepart}{1}

\providecommand{\tracestep}[2]{%
  \par\vspace{3pt}
  \noindent{\setlength{\fboxsep}{1.5pt}\colorbox{WeChatGreen}{\textcolor{white}{\textbf{\,Step #1\,}}}}\enspace\textbf{#2}\par\nobreak
  \vspace{1pt}{\color{WeChatLine}\hrule height 0.4pt}\nobreak\vspace{1pt}
}
\providecommand{\casecaption}[1]{%
  \par\noindent\begin{minipage}{\linewidth}\captionof{figure}[]{#1}\end{minipage}\par
}
\providecommand{\casefilm}[1]{%
  \par\noindent\includegraphics[width=\linewidth]{figures/cases/#1_film.pdf}\par\nobreak\vspace{2pt}
}
\providecommand{\casebreak}{\par\bigskip\Needspace{0.45\textheight}}
\providecommand{\tracefield}[1]{%
  \par\noindent{\color{tracegreen}\textbf{#1.}}\enspace
}

\ifx\tracebox\undefined
\newtcolorbox{tracebox}[3][]{%
  enhanced,
  breakable,
  colback=tracebackground,
  colframe=#3,
  coltitle=white,
  title={#2},
  fonttitle=\footnotesize\bfseries,
  fontupper=\footnotesize,
  boxrule=0.8pt,
  arc=1.5mm,
  left=2.5mm,
  right=2.5mm,
  top=1mm,
  bottom=1.5mm,
  before skip=4pt,
  after skip=4pt,
  before upper={%
    \raggedright
    \setlength{\parindent}{0pt}%
    \setlength{\parskip}{2pt}%
  },
  #1,
}
\newtcolorbox{tracereturn}{%
  enhanced, colback=WeChatPaper, colframe=WeChatGreen, sharp corners,
  boxrule=0pt, leftrule=2pt,
  left=6pt, right=4pt, top=2pt, bottom=2pt, before skip=3pt, after skip=2pt,
  fontupper=\footnotesize,
  before upper={\raggedright\setlength{\parindent}{0pt}{\color{tracegreen}\textbf{Sub-agent returns.}}\enspace},
}
\newtcolorbox{traceverdict}{%
  enhanced, colback=WeChatMint, colframe=tracegreen, sharp corners,
  boxrule=0pt, leftrule=2pt,
  left=6pt, right=4pt, top=2pt, bottom=2pt, before skip=3pt, after skip=0pt,
  fontupper=\footnotesize,
  before upper={\raggedright\setlength{\parindent}{0pt}{\color{tracegreen}\textbf{Verdict.}}\enspace},
}
\fi

\ifnum\casepart=1
\Needspace{10\baselineskip}
\subsection{Qualitative Example of the Inspection Process}
\label{sec:append_cases}
Figures~\ref{fig:case_hailuo_boxes}--\ref{fig:case_sleigh_plane} show complete interactions of the final \methodname{} checkpoint on three generated videos: the main agent's reasoning and chosen inspection (scope in seconds), the response of the invoked sub-agent, and the final verdict. Text is reproduced verbatim from the model outputs, with only formatting changes. The three examples are decided by OCR, audio, and commonsense and physical evidence, respectively, and Appendix~\ref{sec:append_more_cases} gives three more, including a real video.

\casebreak
\casefilm{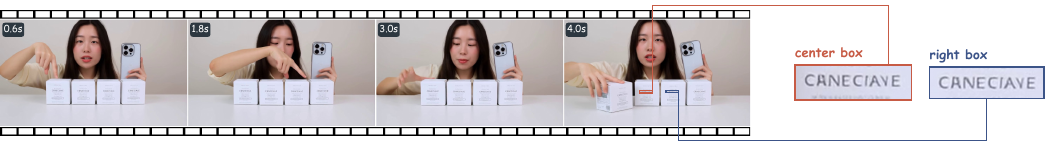}
\begin{tracebox}[unbreakable]{Investigation trace: repeated glyph distortion}{tracegreen}
\sloppy
\tracestep{1}{Main agent \hfill \texttt{visual\_appearance[0.0--5.9 s]}}
\tracefield{Reasoning}
The primary hypothesis is that the sequence may contain a localized temporal rendering defect rather than a globally unstable scene. The overview shows one apparent speaker, a phone, four product boxes, and a hand moving across the foreground, so the most useful first check is a full-range visual comparison across the entire 5.875-second interval. This can establish whether object boundaries, textures, lighting, and motion remain coherent as the hand approaches and leaves the boxes. I will therefore inspect the complete range with the visual-appearance skill, using the sampled progression rather than relying on a single frame. The result should distinguish ordinary motion blur or occlusion from abrupt changes in an object's shape or surface detail. At this stage, the repeated packaging and stable background provide useful reference points, while the moving hand remains the main area requiring scrutiny. Any anomaly found here would motivate a more targeted examination of the visible printed text.
\begin{tracereturn}
\textcolor{flagred}{\textbf{issue} (medium)}. The clearest appearance concern is the unusually smooth, rendered quality of the woman's face during the opening pointing gesture. Across the changing head positions, the forehead, cheeks, and nose retain broad, nearly textureless tonal gradients, while the eyes, eyebrows, and lip contours remain comparatively crisp. The cheek coloration also appears softly blended into the surrounding skin rather than showing comparable fine surface detail. This produces a waxy, airbrushed appearance that persists as the face moves, rather than being confined to one blurred pose. The effect is most apparent when the face is exposed between the moving arm and the phone. Confidence is medium because cosmetic smoothing and image processing can produce similar cues, and the frames do not establish a separate, unmistakable boundary failure.
\end{tracereturn}

\tracestep{2}{Main agent \hfill \texttt{ocr\_consistency[@ 4.0 s]}}
\tracefield{Reasoning}
The preceding full-range appearance review identifies a medium-confidence concern involving the face's unusually smooth, rendered-looking surface, but it does not establish a decisive boundary failure and leaves cosmetic processing as a plausible alternative. The remaining useful question is whether the prominent product packaging behaves like stable physical text. The overview shows four nearly identical boxes with repeated lettering, making a focused OCR-consistency check at 4.0 seconds appropriate: it can compare the legibility and structure of the printed regions while the boxes are clearly visible. This targeted scope is useful because generated imagery often produces malformed or inconsistent glyphs even when the surrounding scene appears coherent. I will use the OCR skill at that timestamp, including its localized text assessment, to test whether the repeated labels retain plausible character shapes and spacing. This evidence should be weighed alongside, rather than substituted for, the earlier appearance finding.
\begin{tracereturn}
\textcolor{flagred}{\textbf{text suspicious} (medium)}. \textit{Issues:} Repeated central A glyphs have unusually narrow upper structures and elongated lower diagonals; The repeated label has inconsistent character construction despite broadly similar lettering. \textit{Reason:} All three crops show a recurring localized anomaly in the central A. The concern is its distorted construction, not the unusual spelling or OCR mismatch. Softness limits certainty, but the repeated structural irregularity supports a suspicious assessment.
\end{tracereturn}

\tracestep{3}{Main agent: final verdict \hfill \textcolor{verdictfake}{\texttt{answer(fake)}}}
\begin{traceverdict}
The evidence supports a fake classification. Across the full sequence, the visual review found a persistent waxy, airbrushed quality in the woman's face: broad textureless gradients coexist with comparatively crisp eyes, brows, and lips. This is a meaningful rendering concern, though its medium confidence and the possibility of cosmetic smoothing or image processing limit its decisiveness. The focused text review independently found recurring structural irregularities in the repeated central ``A'' glyphs, including narrow upper forms and elongated lower diagonals, with inconsistent character construction across similar labels. That result is also medium confidence because softness can obscure fine detail, but the anomaly recurs across multiple localized regions rather than being a single ambiguous crop. Together, the persistent facial rendering artifact and repeated malformed packaging text provide converging signs of synthetic image generation. No audio is available, so there is no speech or voice evidence to weigh. The combined visual and text inconsistencies outweigh the stated uncertainty and favor fake.
\end{traceverdict}
\end{tracebox}
\vspace{4pt}
\casecaption{\textbf{Repeated glyph distortion across product boxes.} Ground truth: fake; \methodname{} prediction: fake; 5.9\,s, silent. After the appearance sub-agent notes a rendered facial surface, an OCR call at 4.0\,s inspects the repeated box labels and finds recurring structural distortion in the central ``A'' glyphs (enlarged on the right).\label{fig:case_hailuo_boxes}}

\casebreak
\casefilm{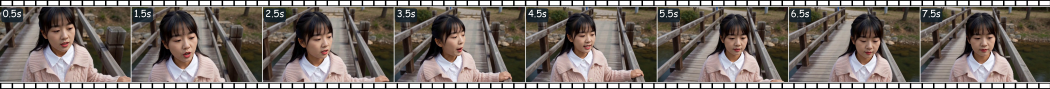}
\begin{tracebox}[unbreakable]{Investigation trace: voice--face mismatch}{tracegreen}
\sloppy
\tracestep{1}{Main agent \hfill \texttt{visual\_appearance[0.0--8.0 s]}}
\tracefield{Reasoning}
The primary hypothesis is that the sequence may contain temporal rendering artifacts rather than merely ordinary camera motion. The supplied overview shows one apparent woman on a wooden bridge across eight seconds, with repeated close views of her face, hair, clothing, and the surrounding railing. A full-range visual appearance inspection is therefore the most useful first step: it can compare identity-bearing details across closely spaced samples while also checking whether the background remains geometrically stable. I will examine the entire 0.0--8.0 second interval rather than a single frame, because generation artifacts often appear as localized changes in facial structure, hair boundaries, or texture that are not visible in isolated images. This scope also allows the tool to distinguish plausible changes caused by pose, focus, lighting, or motion from abrupt inconsistencies that persist despite a largely continuous scene. The result should establish whether there is a concrete visual anomaly before considering the audio evidence.
\begin{tracereturn}
\textcolor{flagred}{\textbf{issue} (medium)}. The face has a conspicuously soft, rendered appearance across the opening portion. The forehead, cheeks, and nose form unusually uniform, low-texture surfaces, while the eyes, eyebrows, and red lip contours remain comparatively crisp. This contrast persists as the head changes angle and the mouth changes shape, giving the facial surface a smoothed, synthetic quality rather than simply making the entire image uniformly soft. The effect is most apparent in the early frontal and slightly turned views, where the surrounding hair strands, ribbed collar, and wooden railing retain more distinct detail. The anomaly is primarily the persistent facial rendering quality, not a definite change in identity or a major boundary failure. Confidence is medium because cosmetic smoothing and image processing can produce some similar characteristics.
\end{tracereturn}

\tracestep{2}{Main agent \hfill \texttt{voice\_profile}}
\tracefield{Reasoning}
The visual inspection has established a medium-confidence concern involving the facial rendering, but that observation alone does not determine whether the media is synthetic because smoothing, compression, or post-processing can create similar effects. The next useful check is the audio, especially because the overview presents one apparent person whose mouth is visibly changing across the sequence. A speaker-aware voice profile can characterize whether the audio contains one coherent speaker and provide descriptive age and gender estimates that may support or conflict with the visible subject. This is not a standalone authenticity test, and the analysis should be discounted if there are multiple speakers, dubbing, voice-over, or weak speaker-to-face correspondence. I will therefore request the profile for the whole video, preserving the full temporal scope so any segmentation or overlap information is available. The purpose is to add an independent modality while keeping the visual anomaly as the main working hypothesis.
\begin{tracereturn}
\textcolor{flagneutral}{\textbf{profile complete} (info)}. \texttt{voice\_profile} (speaker-aware and descriptive; \texttt{profile\_complete} means the analysis succeeded, not that the media is real or fake; emotion dimensions and category probabilities are on 0--1 scales): \texttt{SPEAKER\_00@0.6-2.1s}: \allowbreak{}arousal=0.63, \allowbreak{}dominance=0.66, \allowbreak{}valence=0.43; \allowbreak{}estimated\_age\_years=45, \allowbreak{}female\_probability=0.00, \allowbreak{}male\_probability=1.00, \allowbreak{}child\_probability=0.00; \texttt{SPEAKER\_00@3.3-4.7s}: \allowbreak{}arousal=0.62, \allowbreak{}dominance=0.64, \allowbreak{}valence=0.67; \allowbreak{}estimated\_age\_years=39, \allowbreak{}female\_probability=0.00, \allowbreak{}male\_probability=1.00, \allowbreak{}child\_probability=0.00; overlap\_seconds: 0.0; mossformer2\_used: False; n\_segments: 2.
\end{tracereturn}

\tracestep{3}{Main agent: final verdict \hfill \textcolor{verdictfake}{\texttt{answer(fake)}}}
\begin{traceverdict}
The evidence favors a fake classification. The full-range visual analysis found a persistent, medium-confidence rendering issue: the face appears unusually soft and uniformly low-textured while the eyes, brows, lips, hair, collar, and railing retain sharper detail. That selective smoothing is consistent with synthetic facial rendering, although cosmetic processing remains a plausible limited alternative, so the visual result is not decisive by itself. The audio profile is formally informational rather than an authenticity verdict, but it reports a single speaker with estimated ages of 45 and 39 and male probabilities of 1.00, while the visible, apparently speaking subject is clearly a young woman. This is a concrete age and gender conflict across both detected segments. Dubbing or uncertain correspondence could weaken it, but the overview shows one unambiguous visible speaker and no reported overlap, making the mismatch meaningful supporting evidence. Together, the visual artifact and strong voice-to-face inconsistency outweigh the stated uncertainty.
\end{traceverdict}
\end{tracebox}
\vspace{4pt}
\casecaption{\textbf{Demographic voice--face mismatch on a wooden bridge.} Ground truth: fake; \methodname{} prediction: fake; 8.0\,s, with audio. The appearance sub-agent reports selective facial smoothing, and the voice-profile tool estimates a male speaker aged 45 and 39 (male probability 1.00) across both speaking intervals although the visible speaker is a young woman.\label{fig:case_audio_petite_mismatch}}

\casebreak
\enlargethispage{3\baselineskip}
\casefilm{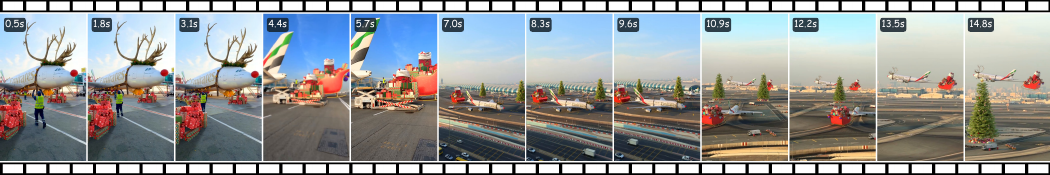}
\begin{tracebox}[unbreakable]{Investigation trace: airborne sleigh}{tracegreen}
\sloppy
\tracestep{1}{Main agent \hfill \texttt{common\_sense[9.4--15.1 s]}}
\tracefield{Reasoning}
The visible sequence presents a highly unusual airport event: a decorated passenger aircraft appears to be moving alongside a large Santa sleigh filled with gifts. Before assessing whether the imagery is physically credible, it is useful to test the scene against ordinary event logic. The key hypothesis is that the apparent flight may be a staged visual effect, a composited sequence, or an AI-generated depiction rather than a genuine recording. This common-sense review should examine whether the aircraft, sleigh, airport infrastructure, and their apparent relationship form a plausible real-world scenario. It should also distinguish merely fantastical subject matter from contradictions in the depicted event, such as unsupported objects, unexplained attachments, or actions that require impossible mechanics. The selected interval covers the transition from the close ground-level views to the wider airborne views, making it appropriate for evaluating the central causal claim without relying only on isolated visual artifacts.
\begin{tracereturn}
\textcolor{flagred}{\textbf{issue} (high)}. The scene depicts a commercial passenger aircraft taking off from an airport runway while carrying a large red sleigh filled with presents and a Santa-like figure. The sleigh is positioned behind the aircraft and appears to travel with it, but no visible mechanism explains how the sleigh is attached, supported, or propelled. The aircraft's ordinary flight behavior is therefore combined with an unexplained passenger-carrying vehicle that has no plausible operational role in the scene. The Christmas trees and decorations fit a festive theme, but they do not provide a causal explanation for the sleigh's movement or the aircraft's unusual load. The event is clearly visible across multiple frames, making the semantic and causal inconsistency sufficiently observable.
\end{tracereturn}

\tracestep{2}{Main agent \hfill \texttt{physical\_consistency[9.4--15.1 s]}}
\tracefield{Reasoning}
The preceding semantic review identifies a strong causal concern: the sleigh appears to accompany the aircraft without an evident explanation for its support or propulsion. That finding supports a more focused physical examination, but it does not by itself establish whether the apparent motion is a practical stunt, an edit, or generated imagery. The next useful test is therefore physical consistency across the same late sequence. This review should track the aircraft's trajectory, the sleigh's position and orientation, the apparent connection between them, and whether their motion remains mechanically coherent as the camera perspective changes. It should also check for continuity in the aircraft's structure and the sleigh's relationship to the surrounding airport environment. Because the available interval is sampled rather than continuous, the assessment should remain attentive to uncertainty about unseen attachment points, forces, and intermediate motion, while still determining whether the visible dynamics provide a credible real-world explanation.
\begin{tracereturn}
\textcolor{flagred}{\textbf{issue} (high)}. The aircraft and sleigh do not maintain physically plausible contact or motion relationships. The sleigh appears to travel alongside and then behind the aircraft while remaining suspended above the tarmac, with no visible support, propulsion, or mechanically credible connection. Across the sequence, the aircraft rises and advances while the sleigh stays at a similar height and follows a separate path, despite the apparent proximity and alignment suggesting a shared movement. The sleigh also passes close to the aircraft without a clear collision response or consistent depth relationship. The reindeer antlers on the aircraft are visible, but they do not provide a plausible mechanism for pulling or carrying the sleigh. These are clear violations of contact, trajectory, and object-motion consistency rather than ordinary camera movement or a normal airport maneuver.
\end{tracereturn}

\tracestep{3}{Main agent: final verdict \hfill \textcolor{verdictfake}{\texttt{answer(fake)}}}
\begin{traceverdict}
Both analyses independently identify high-confidence problems in the central event. The common-sense assessment finds that a commercial passenger aircraft appears to take off while carrying or traveling with a large red sleigh and Santa-like figure, but no visible mechanism explains how the sleigh is supported, attached, or propelled. The physical-consistency assessment reinforces this: the sleigh follows a separate path while remaining at a similar height, lacks credible contact or motion relationships, and passes near the aircraft without a consistent depth or collision response. The antlers are decorative and do not supply a plausible pulling mechanism. The festive decorations and airport setting are coherent in isolation, but they do not resolve the impossible load and motion. Although sampled frames leave some intermediate behavior unobserved, the repeated causal and physical contradictions across the interval are substantial. The media is therefore best classified as fake.
\end{traceverdict}
\end{tracebox}
\vspace{4pt}
\casecaption{\textbf{Passenger aircraft and airborne sleigh in a generated airport scene.} Ground truth: fake; \methodname{} prediction: fake; 16.0\,s, with audio. Commonsense and physical sub-agents both flag the unsupported airborne sleigh and its decoupled trajectory over $[9.4, 15.1]$\,s.\label{fig:case_sleigh_plane}}
\fi

\ifnum\casepart=2
\section{Additional Inspection Examples}
\label{sec:append_more_cases}
Figures~\ref{fig:case_serum_ocr}--\ref{fig:case_street} give three further complete investigations in the format of Section~\ref{sec:append_cases}, covering appearance, physical, commonsense, and OCR evidence on generated videos and a real recording.

\casebreak
\casefilm{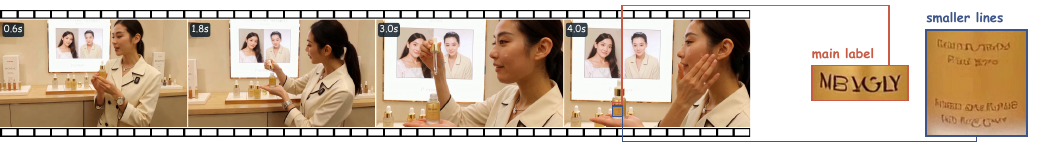}
\begin{tracebox}{Investigation trace: garbled product label}{tracegreen}
\sloppy
\tracestep{1}{Main agent \hfill \texttt{visual\_appearance[0.0--5.0 s]}}
\tracefield{Reasoning}
The sequence shows one apparent presenter demonstrating a cosmetic bottle in a retail-like setting, with the camera moving from a wider view toward a close view. The main hypothesis is that the media may be synthetic if fine details fail to remain stable during this motion, especially on the bottle, dropper, hands, face, and background display. A full-range visual appearance inspection is therefore the most useful first step: it can compare object boundaries, textures, and local details across the entire five-second interval rather than relying on a single frame. This scope also helps distinguish ordinary motion blur or changing perspective from details that appear to regenerate or deform. The result should establish whether there is a broad temporal appearance concern and identify any specific regions that warrant more targeted physical or text-focused examination.
\begin{tracereturn}
\textcolor{flagred}{\textbf{issue} (medium)}. The clearest appearance concern is the unusually smooth, rendered quality of the woman's face during the close-up application sequence. Across the cheek, nose, and jaw, the skin has broad, nearly textureless tonal gradients, while the eyelashes, lip outline, ear, and nearby bottle edges remain comparatively crisp. The cheek droplets also appear as clean, rounded highlights against this uniformly softened surface. This combination gives the face a waxy, synthetic appearance rather than simply making the entire image soft. The effect persists as her hand moves across the cheek, making it more than an isolated blurred view. Confidence is medium because cosmetic smoothing and image processing can produce similar cues, and the frames do not establish a separate, unmistakable boundary failure.
\end{tracereturn}

\tracestep{2}{Main agent \hfill \texttt{physical\_consistency[0.0--5.0 s]}}
\tracefield{Reasoning}
The preceding appearance review raises a possible synthetic-rendering concern, but that concern is not by itself decisive because cosmetic smoothing, lighting, and compression can create similar visual effects. The next question is whether the depicted action obeys ordinary physical continuity. The presenter handles a bottle, raises a dropper, and brings liquid toward her face, so the full five-second interval should be checked for stable hand anatomy, plausible contact, consistent object motion, and coherent liquid behavior. A physical-consistency analysis over the same range is useful because it can either provide independent support for the appearance concern or reveal that the motion remains plausible despite visual stylization. The available one-second overview is sparse, so this check should be interpreted with appropriate caution, particularly around the exact moment of dropper contact and any brief transitions between sampled frames.
\begin{tracereturn}
\textcolor{flagred}{\textbf{issue} (medium)}. The woman holds a dropper above an open bottle and then brings it toward her face. In the close views, the dropper tip and the small amount of liquid on her cheek are visible, but the frames do not clearly show a continuous, mechanically clear transfer from the dropper to the skin. The dropper is held upright with its tip near the bottle opening in the earlier close views, then appears separated from the bottle as she raises it; the later cheek contact is shown in a different pose. This makes the liquid's path and the precise contact relationship difficult to follow. The woman's hand and arm movements otherwise look broadly plausible, and the frames do not establish a definite structural deformation or impossible body motion. The concern is limited to the visible dropper-to-skin interaction, where continuity and contact are not convincingly demonstrated.
\end{tracereturn}

\tracestep{3}{Main agent \hfill \texttt{ocr\_consistency[@ 4.0 s]}}
\tracefield{Reasoning}
The two preceding full-range checks leave the media's printed and displayed lettering as an important unresolved area. The scene contains several product packages and a large background sign, and generated imagery often produces locally unstable or malformed glyphs even when the overall composition looks coherent. A targeted OCR consistency check at 4.0 seconds is appropriate because the close-up composition makes the foreground bottle and nearby text more legible than in the opening wide view. This call should compare the apparent lettering across localized regions and determine whether any irregularity is confined to low-resolution text or affects prominent, readable labels. Such evidence is supporting rather than standalone proof: perspective, focus, motion, and stylized branding can reduce recognition quality. Nevertheless, a high-confidence text anomaly in a clear frame would materially strengthen the synthetic-media hypothesis when combined with the earlier temporal appearance and physical-continuity findings.
\begin{tracereturn}
\textcolor{flagred}{\textbf{text suspicious} (high)}. \textit{Issues:} Malformed and irregularly branching strokes in the prominent bottle label; Fused strokes and unstable character boundaries across multiple smaller label lines; Structurally unresolved lettering on both foreground and background bottles. \textit{Reason:} The supplied crops show repeated glyph-structure defects across separate text regions. The prominent label provides the clearest evidence, while the smaller lines and background label show additional supporting degradation. The assessment rests on visible stroke formation, not on the OCR strings or the wording they represent.
\end{tracereturn}

\tracestep{4}{Main agent: final verdict \hfill \textcolor{verdictfake}{\texttt{answer(fake)}}}
\begin{traceverdict}
The evidence converges on a synthetic-media classification. The full-range visual inspection identifies a persistent waxy, unusually smooth facial surface with crisp nearby details, which is a medium-confidence appearance concern rather than an unmistakable boundary failure. The physical review independently finds a medium-confidence issue in the dropper-to-cheek interaction: the sampled poses do not clearly establish a continuous liquid transfer, although the broader hand and arm motion remains plausible. That limitation means the physical result is supportive, not decisive. The strongest evidence comes from the targeted text analysis at 4.0 seconds, which reports high-confidence malformed, fused, and structurally unresolved lettering across multiple foreground and background regions, including the prominent bottle label. These defects are not merely uncertain OCR readings; they concern visible glyph formation. Taken together, the repeated text irregularities, persistent rendered facial appearance, and unclear liquid contact outweigh the otherwise plausible staging and motion. The appropriate classification is fake.
\end{traceverdict}
\end{tracebox}
\vspace{4pt}
\casecaption{\textbf{Garbled product label in a generated cosmetics demonstration.} Ground truth: fake; \methodname{} prediction: fake; 5.0\,s, silent. The appearance and physical sub-agents identify a rendered facial surface and an unconvincing dropper-to-cheek liquid transfer, and the main agent targets the bottle with an OCR call at 4.0\,s, which finds malformed strokes in the main label and fused characters in the smaller lines. On the 4.0\,s frame, colored boxes mark the PP-OCR text lines of the main label and the smaller lines, enlarged on the right.\label{fig:case_serum_ocr}}

\casebreak
\casefilm{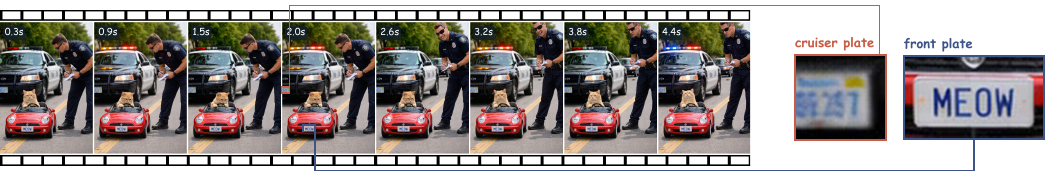}
\begin{tracebox}{Investigation trace: cat at the wheel}{tracegreen}
\sloppy
\tracestep{1}{Main agent \hfill \texttt{common\_sense[0.0--5.0 s]}}
\tracefield{Reasoning}
The sequence presents a highly unusual staged-looking interaction: a cat appears positioned at the steering wheel of a small red car, while an officer stands beside it and a police vehicle is behind them. Before assessing rendering artifacts, the first question is whether the depicted action is physically and semantically plausible. A common-sense review across the full five-second interval can test whether the cat's posture, apparent control of the vehicle, officer's behavior, and vehicle motion form a coherent real-world event. This broad temporal scope is useful because a single frame could make an unusual but possible arrangement seem credible, whereas repeated frames may reveal whether the cat actually interacts with the controls or whether the scene remains frozen. The result should establish whether the central premise itself is credible, while leaving image-generation artifacts and text quality for separate checks.
\begin{tracereturn}
\textcolor{flagred}{\textbf{issue} (high)}. The scene presents a cat as the apparent driver of a full-size road car, seated upright behind the steering wheel and apparently operating it while a police officer writes a ticket beside the vehicle. The cat's paws are placed on the wheel, but there is no visible human operator, restraint, or other mechanism that would make the cat's control of the vehicle plausible. The officer's ticketing behavior is coherent on its own, yet the central event depends on an animal performing a human driving task without a credible causal explanation. The car remains stationary in the sampled frames, so the anomaly is not based on an observed crash or impossible movement; it is the depicted affordance and action itself. The scene therefore conflicts with ordinary animal capabilities and vehicle operation.
\end{tracereturn}

\tracestep{2}{Main agent \hfill \texttt{visual\_appearance[0.0--5.0 s]}}
\tracefield{Reasoning}
The preceding assessment identifies a strong plausibility concern centered on the cat's apparent role as driver, but that concern does not by itself distinguish a deliberately staged scene from synthetic imagery. The next useful test is temporal visual consistency across the same complete interval. Comparing the cat's face, paws, vehicle edges, officer, police car, and background can reveal whether details remain stable as the officer changes pose and the apparent vehicle motion progresses. Particular attention should go to fine boundaries and repeated structures, since generated imagery often preserves the broad composition while subtly regenerating local features between frames. This check can either provide independent support for an artificial origin or show that the unusual premise is visually coherent. It should remain focused on appearance continuity rather than re-evaluating the semantic plausibility already examined.
\begin{tracereturn}
\textcolor{flagred}{\textbf{issue} (medium)}. The scene has a conspicuously smooth, rendered appearance, most evident in the cat's face and the officer's exposed skin. Across the opening views, the cat's forehead, cheeks, and muzzle have broad, softly blended fur shading with little fine strand detail, while its eyes and facial outline remain unusually clean and graphic. The officer's cheek, forehead, and forearms similarly show smooth tonal transitions and simplified surface detail. This treatment contrasts with the more sharply delineated road texture, vehicle edges, and clothing outlines. The concern is the persistent synthetic softness of the subjects, rather than a definite change in identity or a lighting contradiction. The visible evidence supports an appearance anomaly with moderate confidence; it does not establish a specific temporal flicker or a particular facial feature changing shape.
\end{tracereturn}

\tracestep{3}{Main agent \hfill \texttt{ocr\_consistency[@ 2.0 s]}}
\tracefield{Reasoning}
The two preceding checks leave the main question focused on whether the scene's visible text behaves like ordinary photographed signage. The license plate is prominent and contains a short, high-contrast word, making it a useful localized test: generated images often produce readable-looking lettering with unstable or malformed character boundaries, while genuine text may remain structurally consistent even when slightly blurred. Inspecting the plate at the specified two-second point is appropriate because it is clearly visible in the supplied chronology and can be evaluated independently of the cat-and-driver premise. This is a narrow corroborating check, not a standalone authenticity test. Its value depends on whether the lettering is legible and whether any irregularity is localized to the text rather than caused by general image softness, perspective, or compression.
\begin{tracereturn}
\textcolor{flagred}{\textbf{text suspicious} (medium)}. \textit{Issues:} Localized background plate lettering is structurally unreadable and appears melted or unresolved; The clear foreground plate provides a strong contrast with the degraded background plate text. \textit{Reason:} One localized text region shows generation-like glyph dissolution and unstable character structure, while the other is regular and legible.
\end{tracereturn}

\tracestep{4}{Main agent: final verdict \hfill \textcolor{verdictfake}{\texttt{answer(fake)}}}
\begin{traceverdict}
The evidence converges on a synthetic or otherwise non-real depiction. The strongest finding is the high-confidence common-sense issue: the cat is presented as the apparent driver of a road vehicle, with paws on the wheel but no credible mechanism explaining how it could operate the car. The officer's ticketing behavior is coherent, but it does not resolve that central physical contradiction. Visual inspection adds moderate support, describing unusually smooth, rendered-looking fur and exposed skin, with simplified shading and clean graphic facial contours compared with sharper environmental details. The text check is also moderately suspicious: one localized plate region has melted or unresolved lettering, while another is clear, creating an inconsistent text-quality pattern. These appearance and text findings are not individually decisive, and the stationary vehicle limits motion-based inference, but together with the implausible central action they strongly favor an AI-generated scene. Therefore the media is fake.
\end{traceverdict}
\end{tracebox}
\vspace{4pt}
\casecaption{\textbf{AI-generated traffic stop with a cat at the wheel.} Ground truth: fake; \methodname{} prediction: fake; 5.0\,s, silent. Commonsense and appearance sub-agents flag the cat-as-driver premise and rendered fur and skin, and an OCR call at 2.0\,s contrasts the legible foreground plate with the dissolved lettering on the background cruiser plate (enlarged on the right).\label{fig:case_cat_car}}

\casebreak
\casefilm{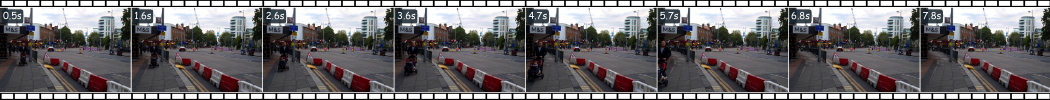}
\begin{tracebox}{Investigation trace: real street recording}{tracegreen}
\sloppy
\tracestep{1}{Main agent \hfill \texttt{visual\_appearance[0.0--8.3 s]}}
\tracefield{Reasoning}
The primary hypothesis is that this is a genuine continuous street recording, but the repeated frames should first be checked for visual artifacts that could indicate synthetic generation. I will examine the full available interval rather than a single image, focusing on whether fixed structures, signage, barriers, vehicles, pedestrians, and the moving foreground subject retain stable geometry and appearance. This broad temporal scope is useful because generated media may reveal inconsistent textures, object boundaries, lighting, or background details even when individual frames look plausible. The foreground man and stroller deserve particular attention because their motion provides a natural test of continuity, while the largely static street environment supplies a reference for detecting unexplained changes. At this stage, the goal is to establish whether there are visible appearance anomalies; physical plausibility and text legibility can be assessed separately afterward.
\begin{tracereturn}
\textcolor{flaggreen}{\textbf{no issue} (medium)}. Across the sequence, the street scene retains consistent appearance: the storefront, road markings, barriers, buildings, trees, and parked vehicles keep their shapes, colors, and relative positions. The moving man and stroller remain visually coherent as they approach the left edge, with no clear boundary melting, duplication, or abrupt change in detail. Fine facial and clothing details are small, and the largely fixed viewpoint limits assessment of texture behavior under substantial motion or changing light. Within the visible regions, however, there is no distinct temporal appearance anomaly.
\end{tracereturn}

\tracestep{2}{Main agent \hfill \texttt{physical\_consistency[0.0--8.3 s]}}
\tracefield{Reasoning}
The appearance review provides a basis for treating the scene as visually stable, but it does not by itself establish that the observed motion follows ordinary physical relationships. The next question is whether the man, stroller, road obstacles, vehicles, and surrounding structures move consistently through time. I will therefore inspect the same full interval with a physical-consistency analysis. This scope is appropriate because it can test continuity of trajectories, contact with the pavement, object persistence, and possible distortions such as teleportation, merging, splitting, or implausible deformation. The scene appears to contain limited motion, so the analysis should distinguish absence of detected violations from strong evidence about dynamics that are not sufficiently exercised. In particular, the stroller's path toward the frame edge and the parked or distant traffic should be compared against the fixed barriers, road markings, and buildings. This step can either add support for authenticity or identify a concrete physical contradiction.
\begin{tracereturn}
\textcolor{flaggreen}{\textbf{no issue} (medium)}. The visible motion is physically coherent across the sampled frames. A man pushes a stroller along the sidewalk from the middle-left toward the left edge, with the stroller maintaining a stable shape, orientation, and contact with the pavement. The child remains seated in the stroller, and the man's posture and hand placement are consistent with the movement. The stroller wheels stay aligned with the sidewalk, and there is no visible teleportation, sudden acceleration, limb distortion, object splitting, or implausible collision. The road barriers, parked vehicles, buildings, and other rigid structures remain fixed in place. The 2 FPS sampling does not show every intermediate movement, so very fine kinematic details cannot be checked, but the observed progression over the full sequence is natural and continuous enough to support no physical-consistency issue.
\end{tracereturn}

\tracestep{3}{Main agent \hfill \texttt{ocr\_consistency[@ 4.0 s]}}
\tracefield{Reasoning}
The temporal and physical checks address motion and scene structure, while visible lettering remains an independent source of evidence. Synthetic imagery can produce unstable, malformed, or inconsistent text, especially on storefronts and road signs. I will inspect the text at the supplied four-second point, where the main signage is clearly visible, using OCR consistency rather than attempting to infer meaning from the scene alone. The useful scope is localized: the large shop sign, speed marking, and nearby directional or warning text can be checked for coherent glyph shapes and ordinary perspective effects. This single-time inspection cannot establish temporal text stability across the entire clip, and small or blurred lettering may remain unresolved, so it should be treated as supporting evidence only. Combined with the earlier full-range checks, however, it can reveal whether prominent readable regions show the characteristic warping or incoherence often associated with generated imagery.
\begin{tracereturn}
\textcolor{flaggreen}{\textbf{text ok} (high)}. All supplied text regions show coherent glyph structure consistent with ordinary photographed signage. Minor softness and stylization are present but do not produce visible warping or generation-like text anomalies.
\end{tracereturn}

\tracestep{4}{Main agent: final verdict \hfill \textcolor{verdictreal}{\texttt{answer(real)}}}
\begin{traceverdict}
The full-range visual assessment found no distinct temporal appearance anomaly: the storefront, road markings, barriers, buildings, trees, and vehicles remain consistent, while the man and stroller retain coherent shapes and detail as they move. The physical-consistency assessment independently found natural, continuous motion, with stable stroller orientation, pavement contact, posture, and hand placement, and no observed teleportation, splitting, limb distortion, or implausible collision. Both findings are medium-confidence and the 2 FPS sampling leaves fine intermediate kinematics unresolved, so they are supportive rather than absolute. The OCR check at four seconds was high-confidence and found coherent glyph structure across localized signage, with only minor softness or stylization. Its single-frame scope limits conclusions about text over time, but it provides no contradictory evidence. Taken together, the stable appearance, plausible motion, and coherent text favor an ordinary recorded scene rather than AI generation.
\end{traceverdict}
\end{tracebox}
\vspace{4pt}
\casecaption{\textbf{Real street recording.} Ground truth: real; \methodname{} prediction: real; 8.3\,s, with audio. Three independent checks return no issue, so the main agent concludes real.\label{fig:case_street}}
\fi

\clearpage
\section{Related Work}
\label{sec:related}
\subsection{AI-Generated Video Detection}
Early detectors treat the task as binary classification over generation traces in images~\citep{wang2020cnn,ojha2023towards,wang2023dire,tan2024rethinking} and videos~\citep{bai2024ai,chen2026demamba,zhang2026physics,interno2026ai,li2026preserving}, outputting a score without verifiable evidence, and benchmarks such as GVD~\citep{bai2024ai}, GenVideo~\citep{chen2026demamba}, GenVidBench~\citep{ni2025genvidbench}, and AIGVDBench~\citep{ma2026your} scale up binary evaluation across generators. Building on MLLM-based image forensics~\citep{jia2024can,ye2025loki,li2025fakebench,xu2025fakeshield,huang2025sida}, recent video detectors pair each verdict with a natural-language account of the observed artifacts. Methods such as BusterX~\citep{wen2025busterx}, BusterX++~\citep{wen2025busterx++}, and VidGuard-R1~\citep{park2026vidguard} train this reasoning with reinforcement learning, while Skyra~\citep{li2026skyra}, VideoVeritas~\citep{tan2026videoveritas}, and VidForensics-M1~\citep{liu2026vidforensics} strengthen it with artifact grounding, perception pretext tasks, and temporal-grounding rewards, respectively. IVY-FAKE~\citep{jiang2026ivy} and Omni-Fake~\citep{li2026omni} provide annotated data for explainable detection, while OmniVL-Guard~\citep{shen2026omnivl} unifies detection and grounding across text, images, and videos. However, these detector-centered formulations do not learn when to request a specialized follow-up inspection or how to adapt it to intermediate observations.

\subsection{Agent-Based Forensic Reasoning}
General video agents query specialized models to gather evidence over long inputs~\citep{wang2024videoagent,xing2026native,li2026omnigaia}, while forensic agents organize analysis through explicit tools or specialized roles. Tool-augmented detectors such as ClueAegis~\citep{cao2026clueaegis}, OmniVL-Guard Pro~\citep{shen2026omnivlpro}, and TUE-Detector~\citep{wu2026tue} train a single MLLM to invoke specific toolchains or visual operations such as cropping, zooming, and segmentation. Multi-agent methods such as SafeGuard~\citep{wu2026safeguard}, Hermes~\citep{lihermes}, and DVAR~\citep{qi2026dvar} instead distribute evidence extraction, verification, and deliberation across specialized agents. ARGUS~\citep{zou2026multi} and Deepfake-Agent~\citep{guo2026deepfake} further train a judge or aggregator over fixed expert agents. However, the invoked experts in these designs never learn from the interactions they serve while the coordinating policy is trained, leaving the experts' forensic response policies outside the ongoing policy-learning loop.

\section{Conclusion}
This paper presents \methodname{}, an agentic framework for AI-generated video detection in which an investigator coordinates a self-evolving library of forensic inspectors. Within this framework, \trainname{} is introduced to optimize the investigator on a fast timescale while refining the inspectors from its calls on a slow one. To support training and evaluation, \datasetname{} is constructed with role-specific supervision and executed trajectories, and \benchname{} is established to assess both detection verdicts and the inspections behind them. Extensive experiments across benchmarks demonstrate the effectiveness of \methodname{}, highlighting the value of co-evolving the investigator with its inspectors.

\clearpage
\bibliographystyle{plainnat}
\setlength{\bibhang}{0pt}
\setlength\bibindent{0pt}
\bibliography{main}

\clearpage
\appendix
\section{Implementation Details}
\label{sec:append_impl}
\paragraph{Training and inference setup.}
The main agent and the four visual sub-agents are rank-64 LoRA adapters on a shared Qwen3.5-4B backbone, initialized from SFT adapters trained on 33{,}896 interaction trajectories and 117{,}552 role-specific records, while the base Qwen3.5-4B serves as the frozen CLUE reference $p_{\mathrm{ref}}$. The main agent observes up to 16 uniformly sampled frames of the full video, and each window-based visual sub-agent inspects its requested window at 2~FPS, whereas the OCR sub-agent inspects the frame at its requested timestamp and the two audio tools analyze the whole soundtrack in Sec.~\ref{sec:pool}. In the fast process, GRPO samples $G{=}4$ rollouts per video with 32 videos per batch and at most 4 sub-agent calls per trajectory for one epoch over the RL pool, and the main-agent learning rate warms up linearly to $1{\times}10^{-5}$. CLUE uses $\epsilon{=}0.1$ and $b{=}1$ in Eq.~\eqref{eq:credit} and call-advantage weight $\lambda{=}0.2$, and the tool-coverage F1 bonus has weight $0.3$. In the slow process, calls with utility below $\kappa{=}{-}0.05$ are relabeled as hedges, and each visual sub-agent's adapter is refreshed every $H{=}20$ main-agent steps on class-balanced batches of 32 grounded calls, with the sub-agent learning rate decayed linearly from $3{\times}10^{-5}$ over training progress, decision-token weight $\omega{=}5$, label-correction weight $\beta{=}1$, and teacher EMA rate $\rho{=}0.001$. The audio tools remain fixed during training.

\paragraph{\benchname{} composition and evaluation metrics.}
\benchname{} comprises $2{,}830$ unique videos in total: $2{,}330$ unique generated videos and $500$ fixed real videos. Each of the five forensic categories (\emph{Visual}, \emph{Physical}, \emph{OCR}, \emph{Common sense}, and \emph{Audio}) contains exactly $500$ generated videos verified to exhibit that category's artifact ($170$ generated videos carry annotations for more than one category, yielding $2{,}330$ unique generated clips across the five $500$-clip category sets). For category $s$, binary detection accuracy ($\text{Acc.}_s$) is evaluated under a balanced 1:1 pairing of its $500$ generated videos and the $500$ real videos ($1{,}000$ videos per category), counting invalid or missing outputs as errors. Sub-agent invocation precision ($\text{Prec.}_s$) measures whether the agent calls the relevant tool for its artifact: computed at the video level over the full pool of generated videos, with $\text{Prec.}_s = |C_s \cap G_s| / |C_s|$ where $C_s$ is the set on which the agent invokes tool $s$ at least once and $G_s$ is the set of videos that call for role $s$. Two label systems are involved. The $500$-clip category membership above only defines the accuracy subsets, and a clip assigned to one category can still exhibit artifacts of other roles. $G_s$ for the four visual roles therefore comes from the dense role annotations of Section~\ref{sec:experiments}, where two cross-family VLM annotators independently mark every role worth inspecting on each generated video and a label-blind judge resolves disagreements. Two videos without a resolved annotation are excluded, leaving $2{,}328$ eligible videos for the visual roles. These annotations do not cover audio, so $G_s$ for audio is the $500$-clip Audio category over all $2{,}330$ generated videos. An agent that invokes tool $s$ on every eligible video attains the base rate $|G_s|/|\text{eligible}|$, which is $91.1\%$, $78.6\%$, $49.2\%$, $24.0\%$, and $21.5\%$ for the appearance, physical, OCR, commonsense, and audio roles. Table~\ref{tab:prec_counts} summarizes these label counts. Precision is read together with category accuracy, which would drop if the agent skipped relevant inspections. \trainname{} raises accuracy on every \benchname{} category (Table~\ref{tab:main_full}), including OCR, commonsense, and audio.

\section{Human Review of Annotations}
\label{sec:human_review}

\begin{figure*}[t]
\centering
\begin{subfigure}[b]{0.625\textwidth}
\centering
\includegraphics[width=\linewidth]{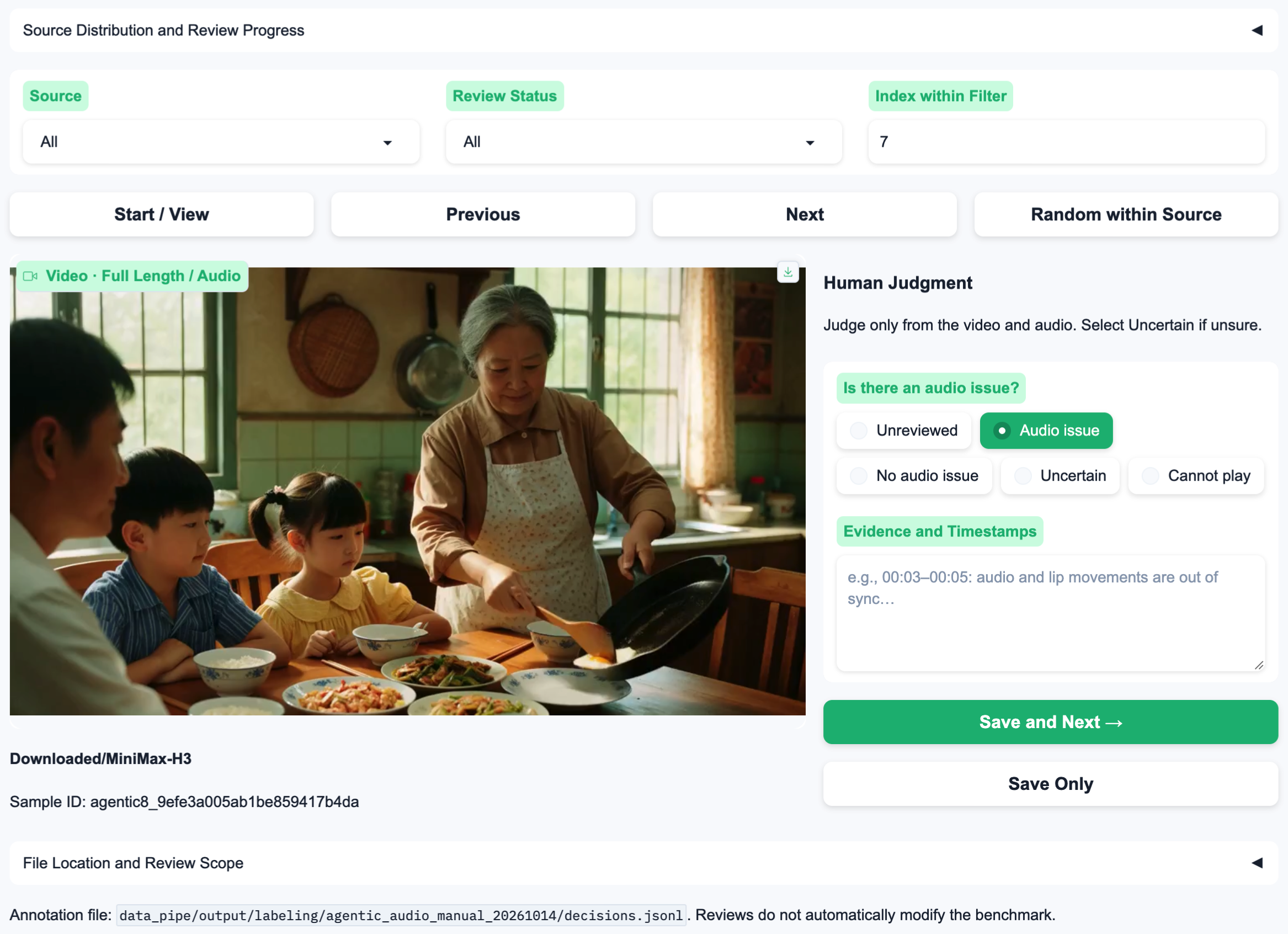}
\caption{Audio-role annotation.}
\label{fig:review_audio}
\end{subfigure}\hfill
\begin{subfigure}[b]{0.34\textwidth}
\centering
\includegraphics[width=\linewidth]{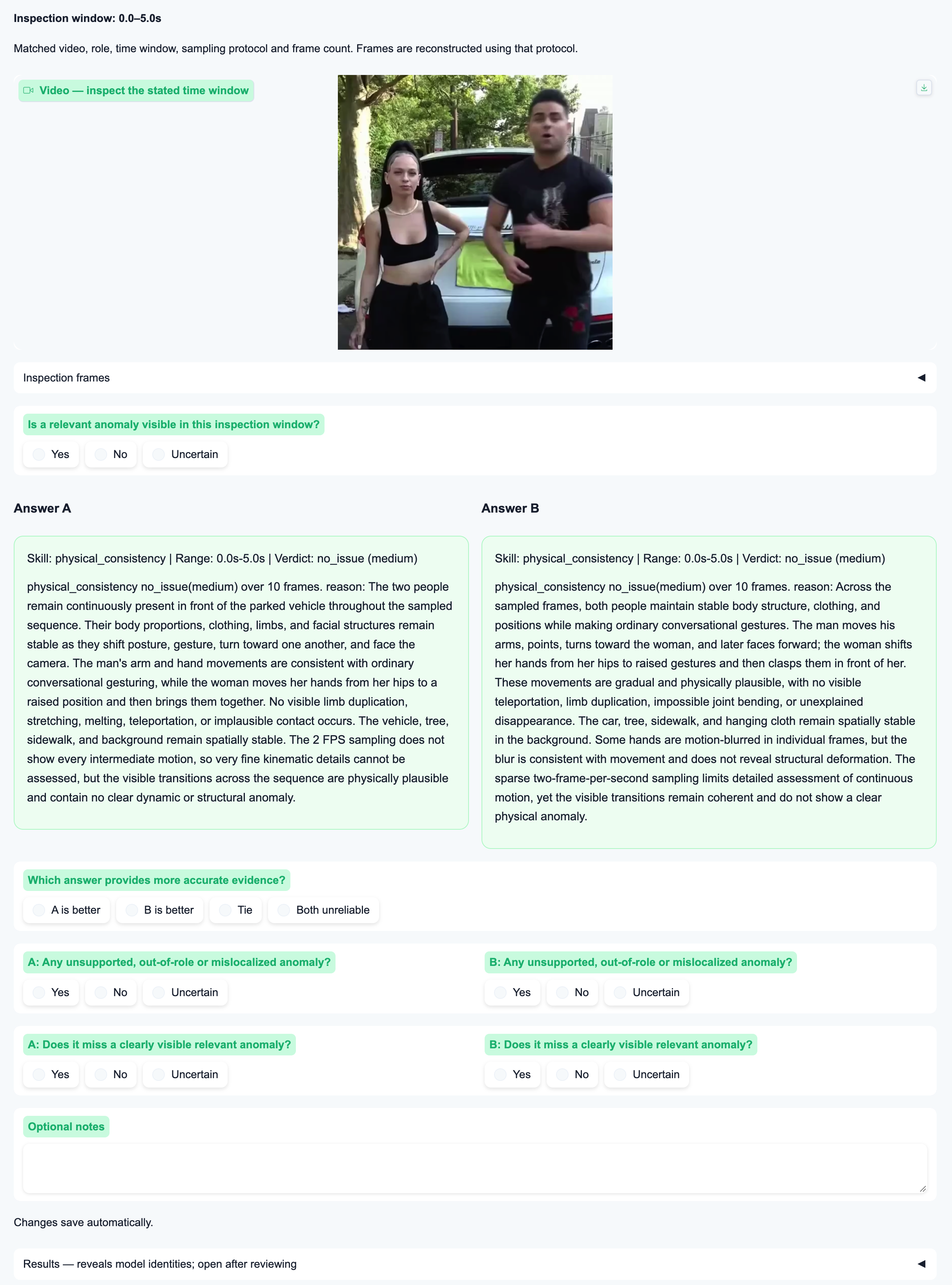}
\caption{Paired sub-agent evidence review.}
\label{fig:review_pair}
\end{subfigure}
\caption{\textbf{Human review interfaces.} \textbf{(a)} The reviewer filters the queue by source and review status, watches the full clip, and records whether an audio inconsistency is present, absent, or inconclusive, along with the supporting time interval. \textbf{(b)} The reviewer inspects the stated window and compares two anonymized sub-agent responses to the same inspection, marking the more accurate evidence and flagging unsupported claims and missed anomalies in each.}
\label{fig:review}
\end{figure*}

Role-specific evidence annotations and main-agent interaction trajectories are produced by model annotators and coupled with human feedback (Figure~\ref{fig:dataset}) before entering the training records of Section~\ref{sec:data}. During prompt development and source onboarding, human reviewers inspect sampled annotations per source and role, flagging hallucinated explanations, over-generalized commonsense claims, imprecise temporal windows, or redundant/omitted tool calls. These failure modes are fed back to refine the annotation prompts and filtering rules, while unplayable clips are discarded. For the audio role, reviewers work through a per-source queue, watch each video with its audio, and record whether the soundtrack is inconsistent with the visible speech along with the offending interval (Figure~\ref{fig:review_audio}).

\begin{figure}[t]
\centering
\begin{subfigure}[t]{162pt}
\centering
\includegraphics[width=158pt]{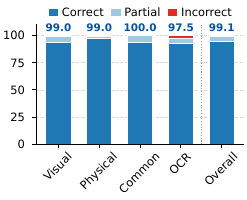}
\caption{Sub-agent annotations.}
\label{fig:human_audit_a}
\end{subfigure}\hspace{8pt}%
\begin{subfigure}[t]{162pt}
\centering
\includegraphics[width=158pt]{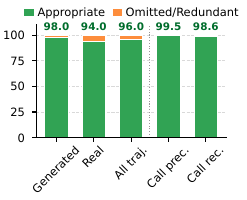}
\caption{Main-agent tool selection.}
\label{fig:human_audit_b}
\end{subfigure}
\caption{\textbf{Quantitative human verification audit of \datasetname{}.} \textbf{(a)} Sub-agent role annotations over $100$ generated videos ($340$ video--role annotations), with numbers above bars giving the at-least-partially-supported rate. \textbf{(b)} Main-agent interaction trajectories over $100$ videos ($50$ generated and $50$ real, $209$ tool calls), with numbers above bars giving trajectory-level tool-selection appropriateness (left of the dotted line) and call-level precision/recall (right).}
\label{fig:human_audit}
\end{figure}

\paragraph{Quantitative human audit of sub-agent annotations and main-agent trajectories.}
To quantify the quality of the resulting supervision in \datasetname{}, we conduct two model-assisted human verification audits (fixed random seed~$42$), summarized in Figure~\ref{fig:human_audit}:
\begin{itemize}
    \item \textbf{Sub-agent role annotations (Figure~\ref{fig:human_audit_a}):} We audit $340$ video--role annotations on $100$ generated training videos across all four visual specialist roles ($100$ \emph{Visual appearance}, $100$ \emph{Physical consistency}, $100$ \emph{Common sense}, and $40$ \emph{OCR consistency} on clips). For each video--role pair, the auditor watches the full video and rates the distilled verdict, temporal interval, and explanation as \emph{Correct \& Supported} (accurately describes a visually observable artifact within the stated window), \emph{Partially Correct} (appropriate verdict with minor explanation or boundary imprecision), or \emph{Incorrect} (hallucinated or contradicted by the video). Overall, \textbf{$99.1\%$ ($337/340$)} of the role annotations are at least partially supported by visual evidence ($94.7\%$ [$322/340$] fully supported and $4.4\%$ [$15/340$] partially correct), with only $0.9\%$ ($3/340$) judged incorrect ($99.0\%$ supported on \emph{Visual}, $99.0\%$ on \emph{Physical}, $100.0\%$ on \emph{Common sense}, and $97.5\%$ on \emph{OCR}).
    \item \textbf{Main-agent tool-selection trajectories (Figure~\ref{fig:human_audit_b}):} We audit $100$ deduplicated main-agent SFT trajectories (comprising $209$ sub-agent invocations across the six forensic roles). For each trajectory, the auditor watches the video with audio, inspects the sequence of invoked tools and returned observations, and judges whether the tool selection is appropriate or suffers from omitted or redundant role invocations. Across the $100$ trajectories, \textbf{$96.0\%$ ($96/100$)} are judged to have appropriate tool selection ($98.0\%$ [$49/50$] on generated videos and $94.0\%$ [$47/50$] on real videos). Only $4$ trajectories exhibit minor selection issues ($3$ omitted calls across \emph{OCR} and \emph{Physical}, and $1$ redundant \emph{OCR} call), corresponding to a call-level precision of \textbf{$99.5\%$ ($208/209$)} and call-level recall of \textbf{$98.6\%$ ($208/211$)}.
\end{itemize}

\paragraph{Paired review of evolved sub-agent evidence.}
Verdict accuracy alone does not show whether evolution improves the evidence a sub-agent reports. We pair $200$ inspections from the evaluation benchmarks, each answered by the SFT and the evolved sub-agent under the same main agent and inspection window. The pairs are stratified by role, video label, and verdict agreement. A reviewer sees both responses in random order with model identities hidden (Figure~\ref{fig:review_pair}), marks which response gives more accurate evidence, and flags in each response unsupported or mislocalized claims and missed visible anomalies. A second reviewer re-checks the judgments. The evolved sub-agents are preferred in $65$ pairs against $20$, and the share of responses with any flaw falls from $30.5\%$ to $8.0\%$ (Table~\ref{tab:paired_review}). Missed anomalies drop from $28.6\%$ to $2.5\%$ and nearly vanish for appearance, physical, and OCR alike, while unsupported claims stay rare. Where the two verdicts agree, neither response is rated better in $97$ of $100$ pairs, so evolution leaves consistent evidence intact and its gains concentrate on the inspections it changes.

\begin{table}[t]
\centering\small
\caption{\textbf{Paired human review of sub-agent evidence.} Each pair answers the same inspection with the SFT (S) and evolved (E) sub-agent. Preferred counts pairs where one response gives more accurate evidence, omitting ties and pairs where both are unreliable. Missed: a clearly visible relevant anomaly is not reported. Flawed: the response misses a visible anomaly or reports one that is unsupported, out of role, or mislocalized. Rates in \%.}
\label{tab:paired_review}
\setlength{\tabcolsep}{4pt}
\begin{tabular}{lrcccc}
\toprule
 & & \multicolumn{2}{c}{Preferred} & Missed & Flawed\\
\cmidrule(lr){3-4}
Sub-agent & Pairs & S & E & S$\rightarrow$E & S$\rightarrow$E\\
\midrule
Appearance & 58 & 6 & \textbf{21} & 32.8$\rightarrow$\textbf{1.7} & 34.5$\rightarrow$\textbf{8.6}\\
Physical & 62 & 6 & \textbf{21} & 27.4$\rightarrow$\textbf{1.6} & 29.0$\rightarrow$\textbf{6.5}\\
OCR & 63 & 7 & \textbf{23} & 33.9$\rightarrow$\textbf{4.8} & 36.5$\rightarrow$\textbf{11.1}\\
Commonsense & 17 & 1 & 0 & 0.0$\rightarrow$0.0 & 0.0$\rightarrow$0.0\\
\midrule
\rowcolor{ForestGreen!10}All & 200 & 20 & \textbf{65} & 28.6$\rightarrow$\textbf{2.5} & 30.5$\rightarrow$\textbf{8.0}\\
\bottomrule
\end{tabular}
\end{table}

\section{Detailed Detection Results}
\label{sec:full_detection_results}

This section expands the main comparisons along two axes: the quality of the evidence returned by individual sub-agents (Table~\ref{tab:skills}), and detection accuracy, fake recall, and F1 across benchmarks and generator categories (Table~\ref{tab:main_full}).

\providecommand{\gu}[1]{\,{\scriptsize\textcolor{ForestGreen}{$\uparrow$#1}}}
\providecommand{\gd}[1]{\,{\scriptsize\textcolor{ForestGreen}{$\downarrow$#1}}}
\providecommand{\ru}[1]{\,{\scriptsize\textcolor{BrickRed}{$\uparrow$#1}}}
\providecommand{\rd}[1]{\,{\scriptsize\textcolor{BrickRed}{$\downarrow$#1}}}

\paragraph{Sub-agent evidence improves across roles.}
Evolving the library improves the evidence returned by every trainable role (Table~\ref{tab:skills}). The physical sub-agent moves the most, with F1 on GB rising from 66.5\% to 97.8\%, and the largest gains of the appearance and OCR sub-agents reach 7.0\% and 11.8\%.

Three patterns explain these gains. First, the physical sub-agent stops hedging. Its inconclusive rate falls on every benchmark, from 26.9\% to 0.3\% on GB, 12.3\% to 0.9\% on GB++, 10.5\% to 1.2\% on MintVid, and 6.4\% to 0.4\% on \benchname{}, and its F1 rises by 9.1--31.3\%. This matches the slow process, which relabels correct but inconclusive low-utility answers as hedges. Second, appearance evidence becomes more decisive. The share of annotated generated calls it flags rises from 96.0\% to 99.8\% on GB and from 66.0\% to 76.2\% on MintVid, and its MintVid F1 rises by 7.0\%. Third, OCR evidence becomes sharper, with F1 rising on every benchmark and most on GB++ by 11.8\%. The gains on the out-of-distribution MintVid, 7.0\% for appearance and 19.2\% for physical, show that the evolved evidence transfers to unseen generators.

\paragraph{Detection improves across evaluated categories.}
Relative to the SFT agent, \trainname{} improves accuracy on all six MintVid categories and all five \benchname{} categories (Table~\ref{tab:main_full}), including an OCR fake-recall increase from 95.0\% to 98.4\%. Fake recall alone can be misleading. Gemini-3.1-Pro and DeepTraceReward label almost every video as generated, reaching 99.0\% and 99.5\% fake recall on GB++ but only 54.3\% and 58.8\% accuracy, whereas \methodname{} keeps 97.2\% fake recall at 97.5\% accuracy. Across the 13 evaluated categories, \methodname{} ranks first in both accuracy and F1 on 12. The largest gains over the SFT agent appear on the unseen MintVid generators, where accuracy rises by 7.5\% on Kling, 6.5\% on Sora2, and 5.1\% on Hailuo.

\begin{table}[H]
\centering\small
\caption{\textbf{GenBuster-Bench breakdown across official splits and generators.} Following the official split organization of BusterX++~\citep{wen2025busterx++}, ID, the eight unseen OOD generators, OOD Mean, and Wild report per-subset forgery detection rate (\%), while All reports overall detection accuracy (\%) across the full benchmark.}
\label{tab:gb_splits}
\resizebox{\textwidth}{!}{%
\begin{tabular}{l*{12}{c}}
\toprule
\multirow{2}{*}{\textbf{Method}} & \multirow{2}{*}{\textbf{ID}} & \multicolumn{9}{c}{\textbf{OOD}} & \multirow{2}{*}{\textbf{Wild}} & \multirow{2}{*}{\textbf{All}}\\
\cmidrule(lr){3-11}
& & Sora & Pika & Gen3 & Luma & WanX & Kling & Jimeng & Vidu & Mean & & \\
\midrule
\rowcolor{tablecategory}
\multicolumn{13}{l}{\textit{Binary Detectors}}\\
NSG-VD~\pub{NeurIPS'25} & 35.0 & 29.5 & 24.0 & 50.0 & 43.0 & 27.3 & 32.0 & 32.0 & 28.0 & 32.3 & 32.7 & 46.4\\
ReStraV~\pub{NeurIPS'25} & 55.8 & 46.5 & 24.0 & 33.0 & 32.0 & 28.0 & 25.0 & 19.0 & 26.0 & 30.7 & 32.0 & 51.9\\
\midrule
\rowcolor{tablecategory}
\multicolumn{13}{l}{\textit{Generic MLLMs}}\\
Qwen3.5-4B & 81.9 & 68.0 & 97.0 & 87.0 & 99.0 & 94.0 & 62.0 & 83.0 & 94.0 & 84.6 & 59.3 & 82.9\\
Qwen3.5-397B & 52.3 & 40.0 & 54.0 & 56.0 & 72.0 & 58.7 & 29.0 & 40.0 & 64.0 & 51.5 & 42.7 & 65.1\\
Qwen3-VL-8B & 29.3 & 5.0 & 15.0 & 19.0 & 39.0 & 15.3 & 6.0 & 5.0 & 16.0 & 14.1 & 8.0 & 44.8\\
Qwen3-Omni-30B-A3B & 41.1 & 11.5 & 26.0 & 38.0 & 48.0 & 30.7 & 15.0 & 20.0 & 28.7 & 25.9 & 24.7 & 50.7\\
Gemini-3.1-Pro & 99.8 & 98.5 & 99.0 & 100.0 & 99.0 & 100.0 & 100.0 & 100.0 & 99.3 & 99.4 & 98.7 & 69.4\\
\midrule
\rowcolor{tablecategory}
\multicolumn{13}{l}{\textit{MLLM-based video forgery detectors}}\\
BusterX++~\pub{NeurIPS'26} & 61.5 & 48.5 & 57.0 & 58.0 & 72.0 & 75.3 & 51.0 & 66.0 & 75.3 & 62.7 & 42.0 & 70.1\\
DeepTraceReward~\pub{COLM'26} & 99.5 & 100.0 & 100.0 & 100.0 & 100.0 & 99.3 & 99.0 & 100.0 & 100.0 & 99.8 & 97.3 & 76.4\\
Skyra-RL~\pub{CVPR'26} & 99.0 & 99.5 & 98.0 & 100.0 & 100.0 & 100.0 & 100.0 & 98.0 & 99.3 & 99.4 & 98.0 & 68.2\\
VideoVeritas~\pub{ICML'26} & 99.6 & 99.0 & 93.0 & 99.0 & 100.0 & 100.0 & 100.0 & 100.0 & 100.0 & 99.0 & 84.7 & \underline{94.3}\\
ARGUS & 81.1 & 89.0 & 98.0 & 90.0 & 100.0 & 95.3 & 91.0 & 93.0 & 91.3 & 93.0 & 76.0 & 86.4\\
\midrule
\rowcolor{tablecategory}
\multicolumn{13}{l}{\textit{Agentic investigation}}\\
\methodname{}-SFT & 94.9 & 88.0 & 90.0 & 88.0 & 99.0 & 96.7 & 83.0 & 95.0 & 94.0 & 91.7 & 78.0 & 93.9\\
\textbf{\methodname{}-\trainname{}} & 99.7 & 100.0 & 99.0 & 99.0 & 100.0 & 100.0 & 99.0 & 100.0 & 99.3 & 99.6 & 95.3 & \textbf{97.7}\\
\bottomrule
\end{tabular}
}
\end{table}

\begin{table}[H]
\centering\small
\caption{\textbf{Detailed detection results.} Accuracy (Acc.), recall (R), and F1 (\%) on GenBuster-Bench (GB), GenBuster-Bench++ (GB++), and the six MintVid General categories, with generated videos as the positive class. MintVid categories follow the official repeated-reference weighting. VideoVeritas MintVid R/F1 are reported in Table~6 of~\citep{tan2026videoveritas}. Best and second-best results in each column are bold and underlined, respectively.}
\label{tab:main_full}
\setlength{\tabcolsep}{3pt}
\resizebox{\textwidth}{!}{%
\begin{tabular}{ll*{14}{c}}
\toprule
\multirow{2}{*}{\textbf{Method}} & \multirow{2}{*}{\textbf{Metric}} & \multirow{2}{*}{\textbf{ID}} & \multicolumn{7}{c}{\textbf{OOD}} & \multicolumn{5}{c}{\textbf{OOD-Agentic}} & \multirow{2}{*}{\textbf{Avg.}}\\
\cmidrule(lr){4-10}\cmidrule(lr){11-15}
& & & GB++ & Hailuo & Jimeng & Kling & Seedance & Sora2 & Wan2.5 & Visual & Physical & OCR & Common & Audio & \\
\midrule
\rowcolor{tablecategory}
\multicolumn{16}{l}{\textit{Binary Detectors}}\\
\multirow{3}{*}{NSG-VD~\pub{NeurIPS'25}} & Acc. & 46.4 & 53.7 & 57.1 & 48.5 & 52.6 & 54.7 & 57.1 & 51.0 & 41.7 & 43.2 & 41.7 & 34.2 & 44.4 & 48.7\\
 & R & 33.6 & 32.3 & 37.7 & 24.0 & 32.1 & 25.1 & 39.4 & 29.1 & 28.8 & 31.8 & 28.8 & 13.8 & 34.2 & --\\
 & F1 & 46.1 & 41.1 & 46.4 & 32.5 & 41.3 & 30.9 & 47.8 & 38.1 & 33.1 & 35.9 & 33.1 & 17.3 & 38.1 & --\\
\midrule
\multirow{3}{*}{ReStraV~\pub{NeurIPS'25}} & Acc. & 51.9 & 66.9 & 59.5 & 56.8 & 55.6 & 64.0 & 50.8 & 54.9 & 55.5 & 54.6 & 51.6 & 47.0 & 51.8 & 57.0\\
 & R & 42.5 & 43.8 & 28.8 & 27.1 & 25.0 & 27.6 & 12.5 & 23.5 & 30.6 & 28.8 & 22.8 & 13.6 & 23.2 & --\\
 & F1 & 54.7 & 56.9 & 41.2 & 39.4 & 36.9 & 38.2 & 20.1 & 35.1 & 40.7 & 38.8 & 32.0 & 20.4 & 32.5 & --\\
\midrule
\rowcolor{tablecategory}
\multicolumn{16}{l}{\textit{Generic MLLMs}}\\
\multirow{3}{*}{Qwen3.5-4B} & Acc. & 82.9 & 73.2 & 63.5 & 56.5 & 65.1 & 73.8 & 57.0 & 73.2 & 73.7 & 71.8 & 83.8 & 94.4 & 58.1 & 74.3\\
 & R & 81.6 & 57.4 & 34.9 & 24.3 & 41.0 & 48.3 & 22.6 & 56.7 & 57.2 & 53.4 & 77.4 & 98.6 & 26.0 & --\\
 & F1 & 86.7 & 68.2 & 48.6 & 36.6 & 55.0 & 59.8 & 34.3 & 68.7 & 68.5 & 65.4 & 82.7 & 94.6 & 38.3 & --\\
\midrule
\multirow{3}{*}{Qwen3.5-397B} & Acc. & 65.1 & 66.7 & 54.7 & 52.0 & 55.8 & 63.0 & 51.4 & 63.4 & 69.2 & 66.7 & 86.0 & 94.5 & 54.1 & 65.7\\
 & R & 51.3 & 38.7 & 19.2 & 17.4 & 25.0 & 24.6 & 13.5 & 39.6 & 47.4 & 42.4 & 81.0 & 98.0 & 17.2 & --\\
 & F1 & 66.7 & 53.7 & 29.5 & 27.3 & 37.0 & 35.0 & 21.6 & 52.9 & 60.6 & 56.0 & 85.3 & 94.7 & 27.3 & --\\
\midrule
\multirow{3}{*}{Qwen3-VL-8B} & Acc. & 44.8 & 55.2 & 52.0 & 50.4 & 50.5 & 63.0 & 52.6 & 55.9 & 61.7 & 58.6 & 72.2 & 91.9 & 47.8 & 55.1\\
 & R & 20.7 & 11.8 & 4.8 & 6.9 & 7.4 & 12.8 & 7.7 & 17.6 & 28.4 & 22.2 & 49.4 & 88.8 & 0.6 & --\\
 & F1 & 33.9 & 20.8 & 9.0 & 12.5 & 13.4 & 21.8 & 14.0 & 29.3 & 42.6 & 34.9 & 64.0 & 91.6 & 1.1 & --\\
\midrule
\multirow{3}{*}{Qwen3-Omni-30B-A3B} & Acc. & 50.7 & 55.2 & 51.7 & 51.9 & 52.9 & 61.4 & 54.6 & 56.3 & 57.5 & 54.8 & 70.6 & 85.6 & 43.1 & 55.8\\
 & R & 32.9 & 21.8 & 17.8 & 19.6 & 21.9 & 24.6 & 22.6 & 28.5 & 42.0 & 36.6 & 68.2 & 98.2 & 13.2 & --\\
 & F1 & 47.6 & 32.7 & 26.7 & 29.6 & 32.6 & 34.0 & 33.1 & 40.4 & 49.7 & 44.7 & 69.9 & 87.2 & 18.8 & --\\
\midrule
\multirow{3}{*}{Gemini-3.1-Pro} & Acc. & 69.4 & 54.3 & 55.1 & 54.4 & 54.6 & 45.3 & 48.6 & 54.9 & 61.8 & 61.6 & 63.4 & 63.4 & 63.1 & 59.6\\
 & R & \textbf{99.5} & \underline{99.0} & \textbf{97.3} & \textbf{96.0} & \textbf{96.0} & \textbf{97.5} & \textbf{87.5} & \textbf{96.6} & 96.4 & 96.0 & \textbf{99.6} & 99.6 & \textbf{99.0} & --\\
 & F1 & 81.6 & 68.4 & 68.1 & 68.5 & 68.7 & 59.0 & 62.9 & 69.0 & 71.6 & 71.4 & 73.1 & 73.1 & 72.8 & --\\
\midrule
\rowcolor{tablecategory}
\multicolumn{16}{l}{\textit{MLLM-based video forgery detectors}}\\
\multirow{3}{*}{BusterX++~\pub{NeurIPS'26}} & Acc. & 70.1 & 79.6 & 70.4 & 71.7 & 82.3 & 71.8 & 58.7 & \textbf{87.3} & 64.4 & 65.1 & 71.1 & 87.1 & 49.9 & 72.7\\
 & R & 60.7 & 68.8 & 45.6 & 47.3 & 73.9 & 47.8 & 13.0 & 86.2 & 35.0 & 36.4 & 48.4 & 80.4 & 6.0 & --\\
 & F1 & 73.5 & 77.1 & 55.6 & 57.2 & 77.1 & 57.7 & 19.8 & \underline{84.5} & 49.6 & 51.1 & 62.6 & 86.2 & 10.7 & --\\
\midrule
\multirow{3}{*}{DeepTraceReward~\pub{COLM'26}} & Acc. & 76.4 & 58.8 & 64.5 & 64.1 & 63.9 & 56.5 & 59.3 & 65.5 & 55.5 & 54.2 & 51.9 & 55.4 & 53.5 & 62.9\\
 & R & \textbf{99.5} & \textbf{99.5} & \underline{92.5} & \underline{92.2} & \underline{91.7} & 89.7 & \underline{84.8} & \underline{94.7} & \textbf{98.6} & 96.0 & 91.4 & 98.4 & 94.6 & --\\
 & F1 & 85.2 & 70.7 & 72.0 & 72.6 & 72.5 & 62.4 & 67.5 & 74.0 & 68.9 & 67.7 & 65.5 & 68.8 & 67.0 & --\\
\midrule
\multirow{3}{*}{Skyra-RL~\pub{CVPR'26}} & Acc. & 68.2 & 55.7 & 51.7 & 51.7 & 53.5 & 63.4 & 50.6 & 59.6 & 83.4 & 83.8 & 74.0 & 86.9 & 68.5 & 64.6\\
 & R & 99.1 & 11.7 & 2.7 & 6.9 & 10.8 & 9.9 & 1.0 & 22.3 & 90.8 & 91.6 & 72.0 & 97.8 & 61.0 & --\\
 & F1 & 81.0 & 20.9 & 5.3 & 12.8 & 19.4 & 17.9 & 2.0 & 36.4 & 84.5 & 85.0 & 73.5 & 88.2 & 65.9 & --\\
\midrule
\multirow{3}{*}{VideoVeritas~\pub{ICML'26}} & Acc. & \underline{94.3} & 91.4 & \underline{85.8} & \underline{80.8} & \underline{86.0} & 86.6 & 67.6 & \underline{86.3} & 72.4 & 75.3 & 76.0 & 91.5 & 69.3 & 86.2\\
 & R & 98.3 & 94.9 & 78.1 & 68.4 & 78.6 & 75.7 & 41.1 & 79.3 & 60.8 & 66.6 & 68.0 & 99.0 & 54.6 & --\\
 & F1 & \underline{95.9} & 91.7 & \underline{84.4} & \underline{78.6} & \underline{85.4} & 82.0 & 55.8 & \textbf{85.8} & 68.8 & 72.9 & 73.9 & 92.1 & 64.0 & --\\
\midrule
\multirow{3}{*}{ARGUS} & Acc. & 86.4 & 78.9 & 61.8 & 56.5 & 65.1 & 67.6 & 56.1 & 73.5 & 73.3 & 71.1 & 80.4 & 90.5 & 64.2 & 76.2\\
 & R & 86.3 & 72.5 & 40.4 & 30.5 & 47.2 & 42.9 & 27.6 & 63.5 & 63.6 & 59.2 & 77.8 & 98.0 & 45.4 & --\\
 & F1 & 89.7 & 77.4 & 51.1 & 42.1 & 58.4 & 51.6 & 38.5 & 71.3 & 70.4 & 67.2 & 79.9 & 91.2 & 55.9 & --\\
\midrule
\rowcolor{tablecategory}
\multicolumn{16}{l}{\textit{Agentic investigation}}\\
\multirow{3}{*}{\methodname{}-SFT} & Acc. & 93.9 & \underline{94.2} & 83.8 & 77.9 & 82.1 & \underline{95.6} & \underline{75.9} & 82.5 & \underline{92.4} & \underline{94.2} & \underline{93.6} & \underline{96.0} & \underline{95.2} & \underline{91.3}\\
 & R & 92.2 & 96.9 & 68.5 & 59.5 & 67.6 & 92.6 & 53.9 & 68.4 & 92.6 & \underline{96.2} & 95.0 & \underline{99.8} & \underline{98.2} & --\\
 & F1 & 95.5 & \underline{95.8} & 81.0 & 74.0 & 80.1 & \underline{95.2} & \underline{69.4} & 80.7 & \underline{92.6} & \underline{94.5} & \underline{93.9} & \underline{96.3} & \underline{95.5} & --\\
\midrule
\multirow{3}{*}{\textbf{\methodname{}-\trainname{}}} & Acc. & \textbf{97.7} & \textbf{97.5} & \textbf{88.9} & \textbf{82.9} & \textbf{89.6} & \textbf{98.6} & \textbf{82.4} & 85.6 & \textbf{97.3} & \textbf{98.0} & \textbf{98.0} & \textbf{98.8} & \textbf{97.0} & \textbf{95.2}\\
 & R & \underline{99.3} & 97.2 & 77.4 & 67.0 & 79.9 & \underline{96.6} & 64.6 & 72.1 & \underline{97.0} & \textbf{98.4} & \underline{98.4} & \textbf{100.0} & 96.4 & --\\
 & F1 & \textbf{98.3} & \textbf{97.4} & \textbf{87.3} & \textbf{80.2} & \textbf{88.9} & \textbf{98.2} & \textbf{78.5} & 83.8 & \textbf{97.3} & \textbf{98.0} & \textbf{98.0} & \textbf{98.8} & \textbf{97.0} & --\\
\bottomrule
\end{tabular}}
\end{table}

\clearpage
\section{Learned Skill Cards}
\label{sec:append_skill_cards}
This section lists the complete skill cards learned by the forensic library across four training runs. Each rule is appended to the user prompt of the corresponding sub-agent alongside the inspection query. \texttt{Report} rules specify generation artifacts to flag, while \texttt{Ignore} rules describe benign camera, motion, or compression effects that should not be mistaken for synthesis errors. The learned rules are local and contrastive. A typical \texttt{Report} rule describes an artifact against an intact reference in the same frame, such as a single fused glyph among sharp neighboring characters or melting facial contours against a sharp background, and the matching \texttt{Ignore} rule names the global degradation that produces a similar look, such as uniform blur or compression across the whole region. Because each edit is mined from calls that the current inspector still judges wrongly, the cards are specific to the failures of that run and differ in concrete content across runs. For example, the physical card of Run 2 targets finger articulation and garment topology, whereas that of Run 3 targets rigid tools sliding over granular surfaces and morphing sculptures.

\tcbset{skillcard/.style={breakable, colback=white, colframe=WeChatDeepGreen, boxrule=0.8pt, arc=1.5mm, fonttitle=\footnotesize, fontupper=\footnotesize, boxsep=1pt, top=4pt, bottom=4pt, left=5pt, right=5pt, before skip=6pt, after skip=6pt}}
\makeatletter
\newcommand{\cardneed}[1]{\par\dimen@\pagegoal\advance\dimen@-\pagetotal\ifdim #1>\dimen@\newpage\fi}
\makeatother
\newcommand{\cardrun}[1]{\cardneed{5\baselineskip}\bigskip\noindent{\small\bfseries #1}\par\nopagebreak\vspace{2pt}\nopagebreak}

\cardrun{Run 1}
\begin{tcolorbox}[skillcard, title={\textbf{Appearance} \hfill 6 rules}]
\textbf{R1.} Report: Surfaces exhibiting broad, continuous color gradients that completely lack fine micro-texture while adjacent objects retain sharp, distinct edges.\par\smallskip
\textbf{R2.} Ignore: Uniform frame-wide softness or compression artifacts that affect the entire scene equally without local texture loss.\par\smallskip
\textbf{R3.} Ignore: Motion blur or softness on moving subjects caused by camera movement or subject velocity.\par\smallskip
\textbf{R4.} Ignore: Smooth skin surfaces resulting from shallow depth of field, distance, or soft natural lighting that lacks harsh specular highlights.\par\smallskip
\textbf{R5.} Report: Facial skin that exhibits a uniform, airbrushed finish lacking pores or micro-texture while adjacent hair and background retain sharp detail.\par\smallskip
\textbf{R6.} Ignore: Silhouetted subjects in strong backlight where facial features are obscured by deep shadow and low contrast.
\end{tcolorbox}
\begin{tcolorbox}[skillcard, title={\textbf{Physical} \hfill 6 rules}]
\textbf{R1.} Report: Rigid objects that change structural shape, length, or thickness during continuous motion or camera panning.\par\smallskip
\textbf{R2.} Ignore: Low-resolution softness or motion blur that makes fine details like fingers or distant rigging indistinct.\par\smallskip
\textbf{R3.} Report: A body or object that exits a frame or vanishes from view without a continuous, physically plausible trajectory.\par\smallskip
\textbf{R4.} Ignore: Soft edges or motion blur on small, fast-moving subjects that make their fine structure indistinct.\par\smallskip
\textbf{R5.} Report: A subject's position or pose that shifts abruptly or discontinuously without a corresponding continuous motion trajectory.\par\smallskip
\textbf{R6.} Ignore: Smooth, mannequin-like continuity of a subject's pose or body during camera movement or panning.
\end{tcolorbox}
\begin{tcolorbox}[skillcard, title={\textbf{OCR} \hfill 6 rules}]
\textbf{R1.} Report: Legible text where a single glyph's strokes fuse into a malformed shape while neighboring glyphs remain structurally intact.\par\smallskip
\textbf{R2.} Ignore: Small, distant, or defocused text where soft edges and low resolution uniformly obscure all stroke details.\par\smallskip
\textbf{R3.} Report: Text where internal counters or stroke junctions melt into illegible blobs despite the surrounding letters staying sharp.\par\smallskip
\textbf{R4.} Ignore: Text where soft edges and low contrast uniformly obscure stroke details across the entire region without localized glyph fusion.\par\smallskip
\textbf{R5.} Report: A glyph whose internal topology is illegible while adjacent characters in the same string remain structurally sharp.\par\smallskip
\textbf{R6.} Ignore: Text on a curved or angled surface where perspective foreshortening or motion blur softens the entire region.
\end{tcolorbox}
\begin{tcolorbox}[skillcard, title={\textbf{Commonsense} \hfill 5 rules}]
\textbf{R1.} Report: Objects that appear, disappear, or deform without a physical cause or visible interaction.\par\smallskip
\textbf{R2.} Report: Animals standing upright or performing human-like coordinated actions without visible support.\par\smallskip
\textbf{R3.} Ignore: Blurry motion, camera shake, or low resolution that preserves the scene's spatial coherence.\par\smallskip
\textbf{R4.} Report: A hand gesture or minor action triggering rapid, large-scale environmental changes like seasonal shifts.\par\smallskip
\textbf{R5.} Ignore: Distant or small body parts that appear indistinct due to camera distance or resolution.
\end{tcolorbox}

\cardrun{Run 2}
\begin{tcolorbox}[skillcard, title={\textbf{Appearance} \hfill 5 rules}]
\textbf{R1.} Report: Skin surfaces that are perfectly matte and textureless, lacking the natural sheen or specular highlights expected from lighting on adjacent hair or fabric.\par\smallskip
\textbf{R2.} Report: Metallic or glass object surfaces that display perfectly clean, band-like highlights lacking micro-scratches or environmental noise.\par\smallskip
\textbf{R3.} Ignore: General motion blur or soft focus on hands and objects during rapid movement or interaction.\par\smallskip
\textbf{R4.} Report: Soft, amorphous organic material that shifts shape or internal texture across frames without matching physical deformation or environmental cues.\par\smallskip
\textbf{R5.} Ignore: Uniformly smooth or airbrushed skin appearance caused by cosmetic filters or heavy compression on a moving subject.
\end{tcolorbox}
\begin{tcolorbox}[skillcard, title={\textbf{Physical} \hfill 5 rules}]
\textbf{R1.} Report: Fingers or hands that melt, overlap, or lack consistent joint articulation during continuous gestures.\par\smallskip
\textbf{R2.} Report: A new structural element or object that emerges from the image boundary or background without a visible path.\par\smallskip
\textbf{R3.} Ignore: Motion blur or soft focus on limbs during fast movement that does not alter underlying limb geometry.\par\smallskip
\textbf{R4.} Ignore: Volumetric smoke or dust that expands, shifts, or dissipates smoothly over time without distorting nearby rigid structures.\par\smallskip
\textbf{R5.} Report: A body or garment silhouette that morphs, truncates, or changes topology across frames without a continuous deformation path.
\end{tcolorbox}
\begin{tcolorbox}[skillcard, title={\textbf{Commonsense} \hfill 5 rules}]
\textbf{R1.} Report: Biological growth or developmental changes that complete in seconds.\par\smallskip
\textbf{R2.} Ignore: Anthropomorphic or fantastical staging where physical object permanence and interaction remain consistent.\par\smallskip
\textbf{R3.} Ignore: Stylized visual effects, explosions, or animated sequences with coherent internal causality.\par\smallskip
\textbf{R4.} Report: A person or animal interacting with a large object or creature without visible physical support or a plausible mechanism.\par\smallskip
\textbf{R5.} Report: A scene where the physical scale of a character is inconsistent with the surrounding environment or other objects.
\end{tcolorbox}

\cardrun{Run 3}
\begin{tcolorbox}[skillcard, title={\textbf{Appearance} \hfill 3 rules}]
\textbf{R1.} Ignore: Soft skin texture on faces caused by shallow depth of field or soft studio lighting.\par\smallskip
\textbf{R2.} Report: Smooth, textureless skin on hands or faces that contrasts with sharp edges on nearby clothing or objects.\par\smallskip
\textbf{R3.} Report: Foreground subjects with unnaturally clean, bright edges against a uniformly blurred or low-detail background.
\end{tcolorbox}
\begin{tcolorbox}[skillcard, title={\textbf{Physical} \hfill 6 rules}]
\textbf{R1.} Ignore: Soft or shifting fabric boundaries where the hem blends with sand, shadows, or low-contrast ground textures.\par\smallskip
\textbf{R2.} Ignore: Apparent scene or subject stillness in aerial or wide shots caused by slow movement or frame sampling.\par\smallskip
\textbf{R3.} Report: Rigid tools or instruments that slide across granular surfaces without producing corresponding material displacement.\par\smallskip
\textbf{R4.} Report: Rigid anatomical forms or sculptures whose contours, limbs, or surface geometry morph or warp across frames.\par\smallskip
\textbf{R5.} Ignore: Smoke, steam, or fire plumes that shift, split, or change color due to wind or turbulence.\par\smallskip
\textbf{R6.} Ignore: Apparent limb elongation or sharp joint angles caused by camera perspective or dynamic posing.
\end{tcolorbox}
\begin{tcolorbox}[skillcard, title={\textbf{OCR} \hfill 6 rules}]
\textbf{R1.} Report: Legible characters whose internal strokes or counters are structurally malformed or inconsistent with standard glyph topology.\par\smallskip
\textbf{R2.} Ignore: Small or distant text that is uniformly blurred by scale or motion but retains stable, recognizable glyph shapes.\par\smallskip
\textbf{R3.} Ignore: Text printed on soft or moving surfaces where strokes appear to waver or distort due to surface deformation.\par\smallskip
\textbf{R4.} Ignore: Text on curved, embossed, or reflective surfaces where strokes appear warped by the underlying geometry.\par\smallskip
\textbf{R5.} Report: Legible text where stroke widths or counter shapes fluctuate irregularly between adjacent characters.\par\smallskip
\textbf{R6.} Report: Legible text where character spacing or stroke junctions are irregularly inconsistent within the same word.
\end{tcolorbox}
\begin{tcolorbox}[skillcard, title={\textbf{Commonsense} \hfill 3 rules}]
\textbf{R1.} Report: A spilled or broken item appearing on the floor without a visible drop, collision, or contact event.\par\smallskip
\textbf{R2.} Report: A person's facial expression or body posture that remains unnaturally static while the surrounding environment changes.\par\smallskip
\textbf{R3.} Ignore: Low-resolution motion blur or chaotic crowd movement where objects remain physically grounded.
\end{tcolorbox}

\cardrun{Run 4}
\begin{tcolorbox}[skillcard, title={\textbf{Appearance} \hfill 5 rules}]
\textbf{R1.} Ignore: Motion blur that softens a moving subject's skin or hair boundaries during fast movement.\par\smallskip
\textbf{R2.} Report: A subject's surface lacking fine texture while adjacent high-frequency details remain sharp.\par\smallskip
\textbf{R3.} Ignore: Global softness or compression artifacts that uniformly reduce detail across the entire frame.\par\smallskip
\textbf{R4.} Ignore: Smooth skin or facial surfaces that lack fine pore detail due to beauty filters or high compression.\par\smallskip
\textbf{R5.} Report: Facial or hand contours that locally melt or duplicate while surrounding background details remain sharp.
\end{tcolorbox}
\begin{tcolorbox}[skillcard, title={\textbf{Commonsense} \hfill 3 rules}]
\textbf{R1.} Report: A character's body or props shifting into a physically incompatible form or scale during a transition.\par\smallskip
\textbf{R2.} Report: A physical object appearing on the ground with a spill or impact mark lacking a visible causal action.\par\smallskip
\textbf{R3.} Ignore: Stylized or game-like environments where objects or animals behave according to internal rules.
\end{tcolorbox}

\section{Additional Evolution Comparisons}
\label{sec:append_qual_comparisons}

\begin{wrapfigure}{r}{0.5\textwidth}
  \centering
  \includegraphics[width=\linewidth]{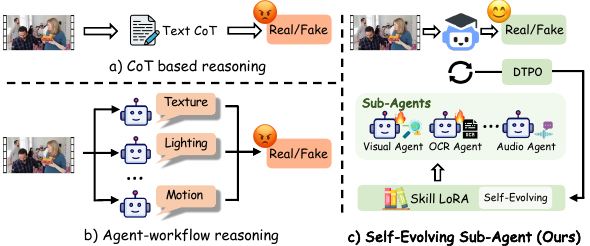}
  \caption{\textbf{Comparison of AI-generated video detection paradigms.} a) CoT-based reasoning uses textual rationales. b) Multi-agent workflows follow fixed execution structures. c) \methodname{} adaptively coordinates sub-agents and evolves them together with the main policy.}
  \label{fig:compare}
\end{wrapfigure}
\begin{figure}[!h]
\centering
\includegraphics[width=0.98\textwidth]{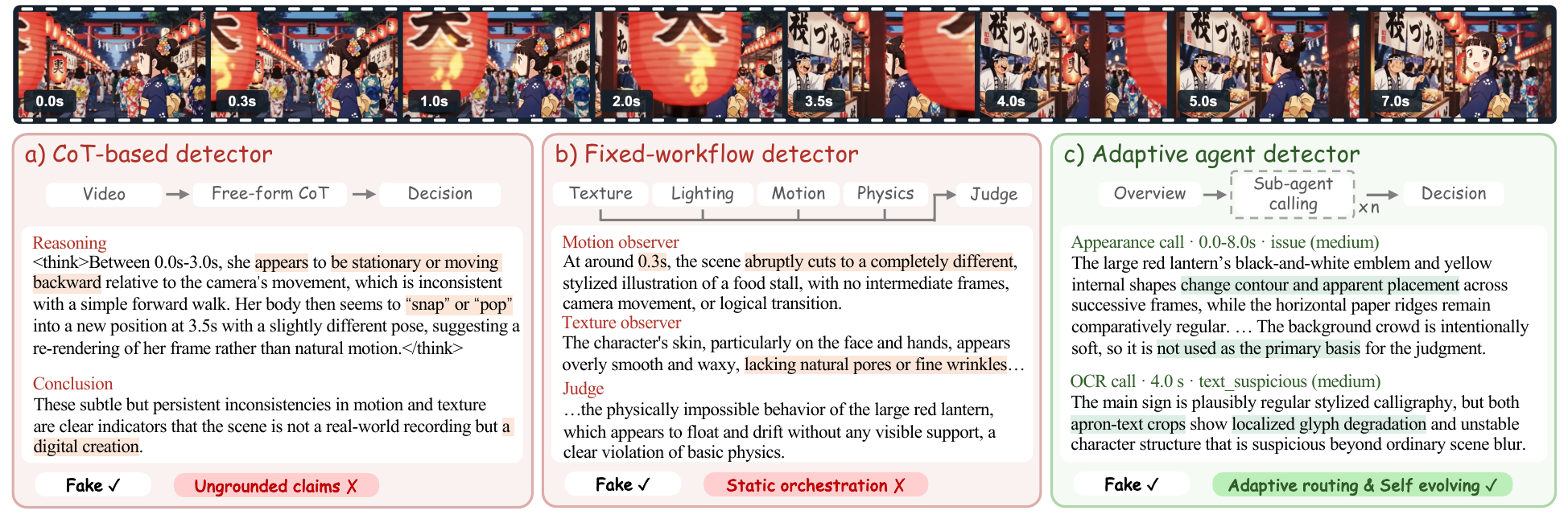}
\caption{\textbf{Qualitative comparison of investigation paradigms.} All three approaches classify the generated video as fake. The CoT-based detector makes unsupported motion claims, and the fixed-workflow detector aggregates predefined observer reports. The adaptive agent requests localized appearance and OCR inspections and grounds its verdict in changing emblem contours and degraded apron lettering.}
\label{fig:motivation_paradigms}
\end{figure}

Figure~\ref{fig:motivation_paradigms} runs the three paradigms of Figure~\ref{fig:compare} on the same generated street-market clip. All three return fake, yet only the adaptive agent grounds its verdict in evidence that exists in the video. The CoT-based detector asserts that the subject moves backward between $0.0$\,s and $3.0$\,s and snaps into a new pose at $3.5$\,s, neither of which happens. The fixed-workflow detector queries four preset observers in parallel, and its judge aggregates reports of an abrupt scene cut at $0.3$\,s and a floating lantern that are equally unsupported. \methodname{} instead issues two targeted calls. An appearance call over $[0.0, 8.0]$\,s finds that the emblem and yellow shapes on the red lantern change contour and placement across frames while its paper ridges stay regular, and it sets aside the soft background crowd as ordinary depth of field. An OCR call at $4.0$\,s then separates the intact calligraphy of the main sign from localized glyph degradation on the apron text.

\begin{figure}[!h]
\centering
\includegraphics[width=0.98\textwidth]{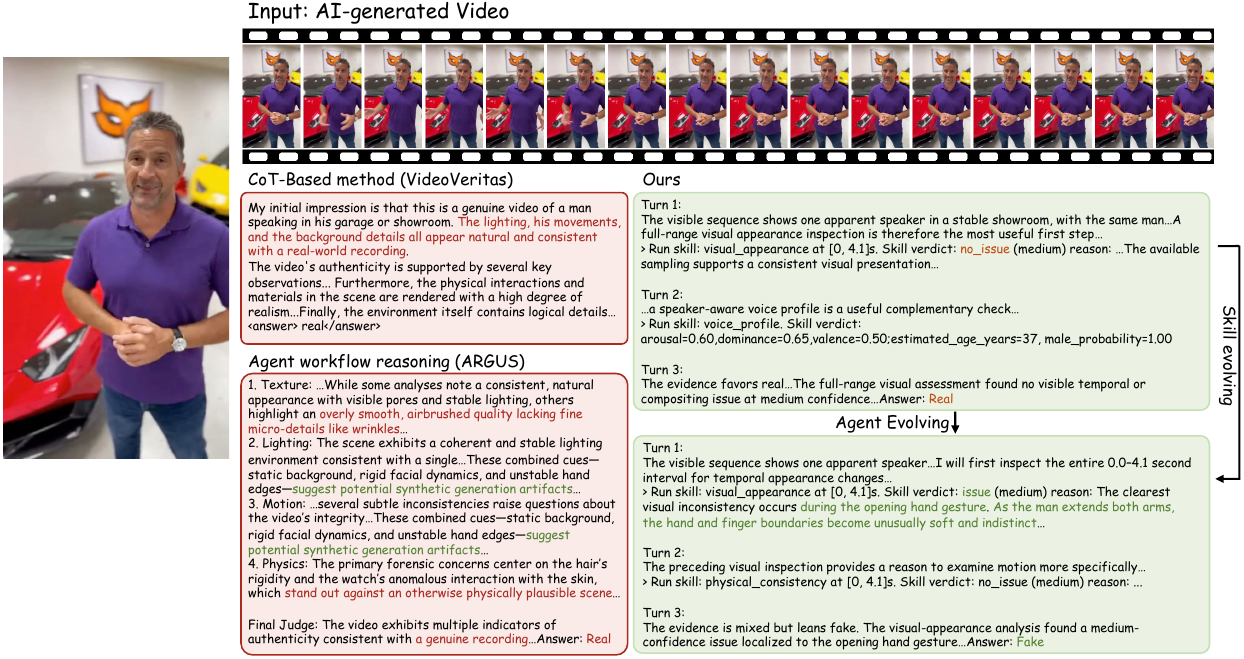}
\caption{\textbf{Complete evolution comparison.} On an AI-generated video, the CoT-based VideoVeritas and the fixed-workflow ARGUS both answer real. Before \trainname{} (top right), the appearance sub-agent reports no issue and \methodname{} answers real. After \trainname{} (bottom right), the same first inspection localizes soft hand and finger boundaries during the opening gesture, and the main agent answers fake.}
\label{fig:visual}
\end{figure}

Figure~\ref{fig:visual} isolates what evolution changes on a generated showroom video. VideoVeritas takes the stable lighting and showroom background at face value and answers real. The ARGUS observers do mention airbrushed skin, unstable hand edges, and rigid hair, but these cues are diluted among generic summaries and the judge still answers real. Before and after \trainname{}, \methodname{} opens with the same appearance call over $[0, 4.1]$\,s. The SFT sub-agent returns \texttt{no\_issue}, a follow-up voice-profile check finds no anomaly, and the agent answers real. The evolved sub-agent returns \texttt{issue} on the identical call, locating soft and indistinct hand and finger boundaries as the speaker extends both arms. The main agent follows up with a physical-consistency check rather than an audio tool and answers fake. Sharper evidence on the same first query thus reverses the verdict and redirects the rest of the investigation.

\par\bigskip

\def\casepart{2}

\end{document}